\documentclass[%
 reprint,
 amsmath,amssymb,
 aps,
]{revtex4-2}

\usepackage{adjustbox}
\usepackage{comment}
\usepackage{graphicx}
\usepackage{dcolumn}
\usepackage{bm}

\usepackage{physics}
\usepackage{xcolor}
\usepackage{hyperref}
\usepackage{multirow}

\newtheorem{definition}{Definition}
\usepackage{color}
\colorlet{Mycolor1}{green!20!orange!80!}
\iftrue  
\newcommand{\comments}[1]{\footnote{\textcolor{blue}{\textit{#1}}}}

\else
\newcommand{\comments}[1]{}
\newcommand{}[1]{#1}
\newcommand{}[1]{#1}
\fi

\begin{document}

\preprint{APS/123-QED}

\title{Separable Hamiltonian Neural Networks}

\author{Zi-Yu Khoo}
\affiliation{School of Computing, National University of Singapore, 13 Computing Drive, Singapore 117417}
\author{Dawen Wu}%
 \email{dawen.wu@cnrsatcreate.sg}
\affiliation{%
CNRS@CREATE, 1 Create Way, 08-01, Singapore 138602
}%
\affiliation{%
School of Computing, National University of Singapore, 13 Computing Drive, Singapore 117417
}%
\author{Jonathan Sze Choong Low}

\affiliation{
Singapore Institute of Manufacturing Technology (SIMTech), 5 Cleantech Loop, 01-01, Singapore 636732
}%

\author{Stéphane Bressan}
\affiliation{%
School of Computing, National University of Singapore, 13 Computing Drive, Singapore 117417
}%
\affiliation{%
CNRS International Research Laboratory 2955 on Artificial Intelligence, Singapore 138632
}%


\date{\today}

\begin{abstract}
Hamiltonian neural networks (HNNs) are state-of-the-art models that regress the vector field of a dynamical system under the learning bias of Hamilton’s equations. 
A recent observation is that embedding a bias regarding the additive separability of the Hamiltonian reduces the regression complexity and improves regression performance.
We propose \textit{separable HNNs} that embed additive separability within HNNs using observational, learning, and inductive biases. We show that the proposed models are more effective than the HNN at regressing the Hamiltonian and the vector field. Consequently, the proposed models predict the dynamics and conserve the total energy of the Hamiltonian system more accurately.  
\end{abstract}

\maketitle


\section{Introduction}
Modeling dynamical systems is a core challenge for science and engineering. The movement of a pendulum, the wave function of a quantum-mechanical system, the movement of fluids around the wing of a plane, the weather patterns under climate change, and the populations forming an ecosystem are spatio-temporal behaviors of physical phenomena described by dynamical systems. Hamiltonian systems~\cite{Meyer1992} are a class of dynamical systems governed by Hamilton's equations, which indicate the conservation of the Hamiltonian value of the system.


Recent advancements in the modeling of Hamiltonian systems include Hamiltonian neural networks (HNNs)~\cite{Marios2022,bertalan_2019,Greydanus_hamiltoniannn_2019}, port-HNNs~\cite{Desai2021} and reservoir computing~\cite{Zhang2021}. Some works also leverage Hamiltonian separability~\cite{Chen_symplectic2020,TothHGN} by using a Leapfrog integrator or requiring the Hamiltonian to be mechanical~\cite{Zhong2019SymplecticOL}. HNNs~\cite{bertalan_2019,Greydanus_hamiltoniannn_2019} use a learning bias~\cite{Karniadakis2021} based on physics information regarding Hamilton's equations~\cite{bertalan_2019,Greydanus_hamiltoniannn_2019} to aid the neural network in converging towards solutions that adhere to physics laws~\cite{Karniadakis2021}. They supervised-ly regress the vector field and the Hamiltonian of a dynamical system from discrete observations of its state space or vector field and outperform their physics-uninformed counterparts in doing so~\cite{khoo_22}. 

Hamiltonian dynamics are often complicated and chaotic, especially in higher dimensional systems such as the Toda Lattice~\cite{Toda1967,Ford1973} and Henon Heiles~\cite{henonheiles1964} systems. A redeeming feature of these systems is their additive separability. As highlighted by Gruver et al., additive separability allows a neural network to avoid artificial complexity from its input variables and improve its performance~\cite{gruver2022deconstructing}. This motivates further informing HNNs of additive separability to alleviate the complexity between variables of Hamiltonian systems.

The main contributions of this work are:
\begin{itemize}
    \item  We propose \textit{separable HNNs} that embed additive separability using three modes of biases: observational bias, learning bias, and inductive bias. 
    \item An observational bias is embedded by training the HNN on newly generated data that embody separability. A learning bias is embedded through the loss function of the HNN. An inductive bias is embedded by conjoined multilayer perceptrons in the HNN. Each proposed separable HNN may embed one or more biases. 
    \item The proposed separable HNNs show high performance in regressing additively separable Hamiltonians and their vector fields. The code for the HNNs is available at \url{github.com/zykhoo/SeparableNNs}.
\end{itemize}

In this paper, Sec.~\ref{sec:background} presents the necessary background on Hamiltonian and separable Hamiltonian systems. Sec.~\ref{sec:methodology} introduces the proposed separable HNNs. Sec.~\ref{sec:exp} compares the proposed HNNs against the baseline HNN in regressing additively separable Hamiltonians and vector fields. Sec.~\ref{sec:timedynamics} compares the proposed HNNs against the baseline HNN on a chaotic system and a high-dimensional system. Sec.~\ref{sec:conclusion} concludes the paper and discusses future work. 

\section{Background} \label{sec:background}

\subsection{Hamiltonian Systems}


A Hamiltonian system is characterized by a Hamiltonian function, $H: \mathbb{R}^{2n} \to \mathbb{R}$. 
The input states to the Hamiltonian function are time-dependent functions $q: \mathbb{R}\to\mathbb{R}^n$ and $p: \mathbb{R}\to\mathbb{R}^n$, conventionally interpreted as position and momentum.
The time derivatives $\dot{q}:= \frac{dq}{dt}$ and $\dot{p}:= \frac{dp}{dt}$ are governed by the following differential equations, called Hamilton's equations:
\begin{eqnarray}
\dot{q} = \frac{\partial H}{\partial p},\quad \dot{p} = -\frac{\partial H}{\partial q}.
\label{eqn:defnHamiltonian2}
\end{eqnarray}
The vector field is the set of time derivatives for all states. 

Eq. \eqref{eqn:defnHamiltonian2} implies that the Hamiltonian is conserved along trajectories as
\begin{equation}\label{eqn:defnHamiltonianconservation}
\frac{dH}{dt} = \frac{\partial H}{\partial q} \cdot \dot{q}  +  \frac{\partial H}{\partial p} \cdot \dot{p} = 0.
\end{equation}


Let $z = (q^T, p^T)^T \in \mathbb{R}^{2\times n}$. For notational simplicity, we omit the superscripts and simplify it to $z = (q, p)$ unless otherwise stated. Let $\bm{J}$ be the $2n \times 2n$ matrix
\begin{eqnarray}
\bm{J} = \left[\begin{array}{cc}
    \bm{0} & \bm{I} \\
    -\bm{I} & \bm{0}
\end{array} \right],
\end{eqnarray}
where $\bm{I} \in \mathbb{R}^{n\times n}$ is the identity matrix and $\bm{0} \in \mathbb{R}^{n\times n}$ is the zero matrix.
Then, Eq.~\eqref{eqn:defnHamiltonian2} can be written compactly as 
\begin{eqnarray}
    \dot{z} = \bm{J} \cdot \nabla_z H,
\end{eqnarray}
where $\nabla_z H = \frac{\partial H}{\partial z}$.


\begin{figure}[h] 
\centering
\includegraphics[width =0.23\textwidth]{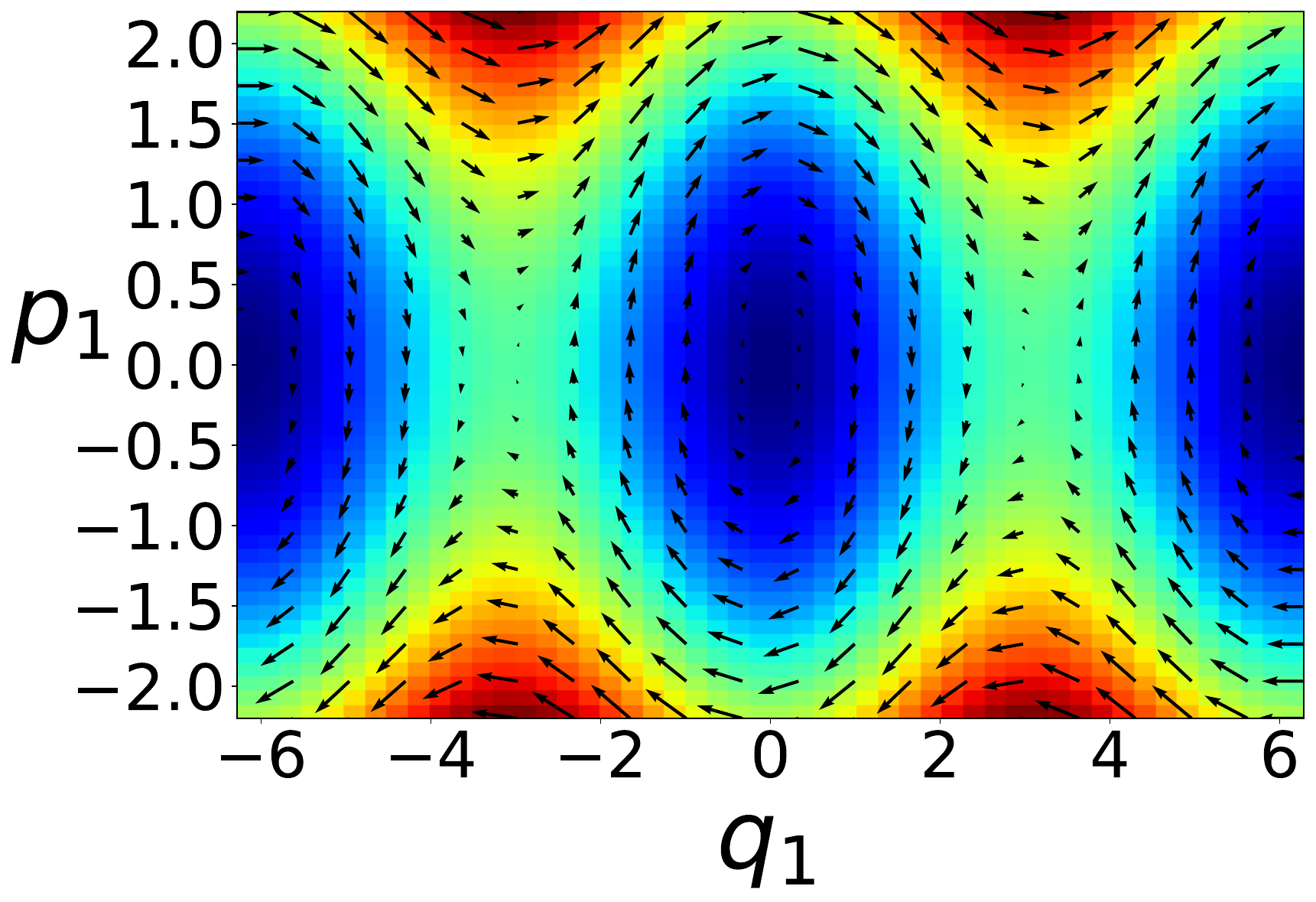} 
\includegraphics[width =0.23\textwidth]{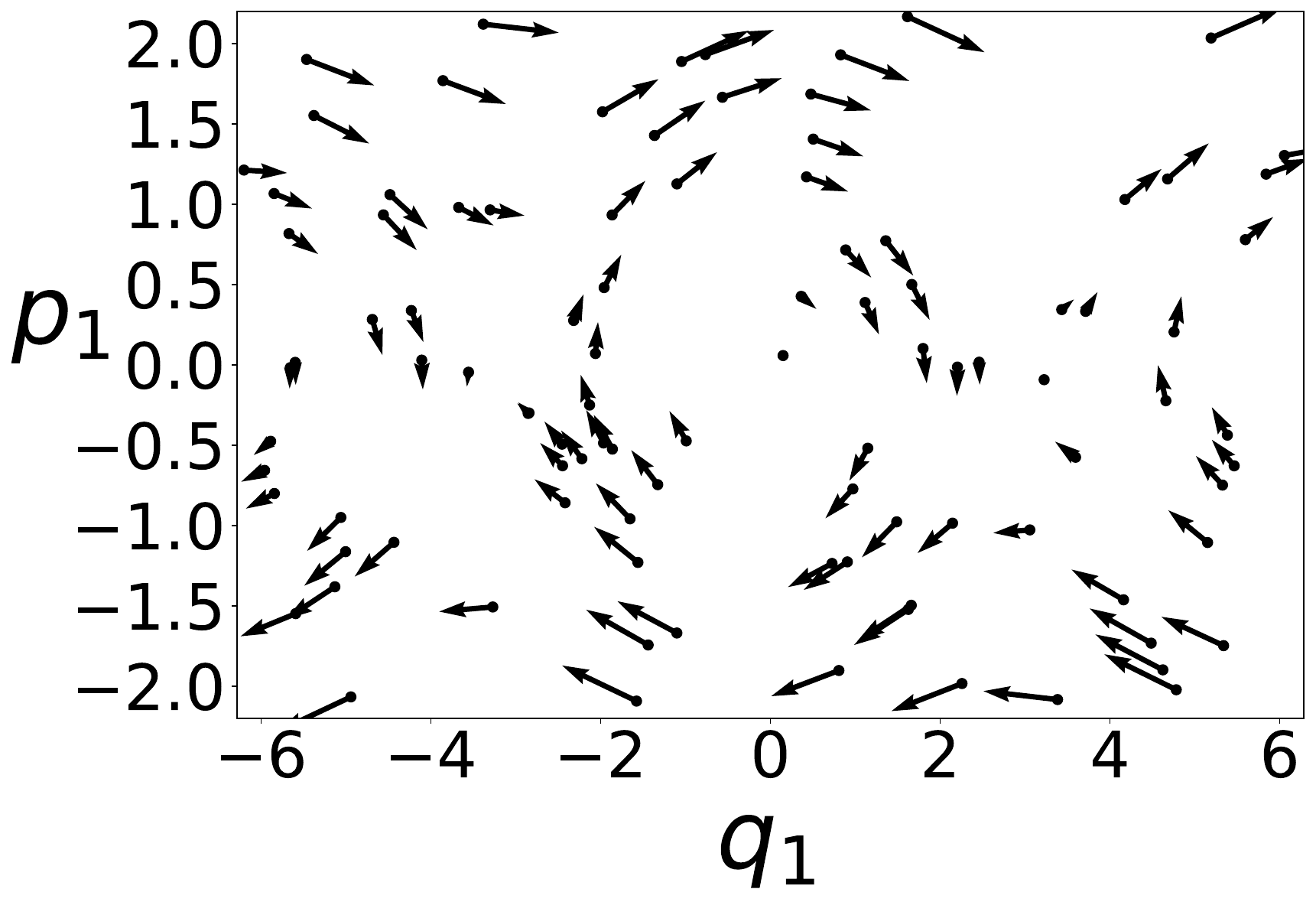} 
       \caption{\label{fig:learning}Left: Nonlinear pendulum vector field (black arrows) and Hamiltonian (heatmap). Right: Random samples $(q, p, \dot{q}, \dot{p})$ of the nonlinear pendulum vector field. }
\end{figure}

A classic example of a Hamiltonian system is the nonlinear pendulum. 
Fig.~\ref{fig:learning} (left) shows the vector field of a nonlinear pendulum in its two-dimensional phase space with states position (angle when oscillating in a plane) and momentum (mass multiplied by velocity) on the x-axis and y-axis respectively. 
For a nonlinear pendulum of unitary mass, the Hamiltonian is the sum of kinetic and potential energy. 
The heatmap in Fig.~\ref{fig:learning} (left) shows the value of the Hamiltonian in the phase space. Fig.~\ref{fig:learning} (right) shows random samples of the same vector field.

\subsection{Separable Hamiltonian Systems}


\begin{definition}[Additive Separability]
    A Hamiltonian system is called additive separable if the Hamiltonian can be separated into additive terms, each dependent on either $q$ or $p$, where $q$ and $p$ are disjoint subsets of the states of a Hamiltonian system. The separable Hamiltonian is defined 
    \begin{eqnarray}
        H(q,p) = T(q) + V(p) \label{eqn:addsep}
    \end{eqnarray}
    where $T: \mathbb{R}^n \to \mathbb{R}$ and $V: \mathbb{R}^n \to \mathbb{R}$ are arbitrary functions. The mixed partial derivative of Eq. \eqref{eqn:addsep} for an additively separable function is necessarily and sufficiently zero~\cite{Udrescu_2020, bellenot_addsep}, i.e.,\begin{eqnarray}\label{eqn:addsep_hamiltonian_mixedpartial}
        \frac{\partial^2 H}{\partial q \partial p}(q,p) =\mathbf{0} . 
    \end{eqnarray}
     for all states $q$ and $p$, and $\mathbf{0}\in \mathbb{R}^{n\times n}$.
\end{definition}

Knowledge regarding additive separability may be embedded within machine learning models as biases. 
Karniadakis et al. present three modes of biasing a regression model: observational bias, learning bias, and inductive bias~\cite{Karniadakis2021}. 
Observational biases are introduced directly through data that embody the underlying physics, or carefully crafted data augmentation procedures. With sufficient data to cover the input domain of a regression task, machine learning methods have demonstrated remarkable power in achieving accurate interpolation between the dots~\cite{Karniadakis2021}. 
Learning biases are soft constraints introduced by appropriate loss functions, constraints and inference algorithms that modulate the training phase of a machine learning model to explicitly favor convergence towards solutions that adhere to the underlying physics~\cite{Karniadakis2021,bertalan_2019,Greydanus_hamiltoniannn_2019}. 
Inductive biases are prior assumptions incorporated by tailored interventions to a machine learning model architecture, so regressions are guaranteed to implicitly and strictly satisfy a set of given physical laws~\cite{Karniadakis2021}.

 \begin{table*}[ht]
\caption{\label{tab:equationsanddimensions}Equations and $n$ of each Hamiltonian system. The subscript represents the index of the element in the state.}
\begin{ruledtabular}
\begin{tabular}{p{4cm}p{12cm}p{0.5cm}}
        System & Equation & $n$ \\
        \hline
        Nonlinear Pendulum & $H(q_1,p_1) = \frac{p_1^2}{2} + (1-\cos{q_1})$ & 1 \\
        Anisotropic Oscillator & $H(q_1,q_2,p_1,p_2) = \sqrt{p_1^2+p_2^2+1} + \frac{q_1^2+q_2^2}{2} + \frac{0\times q_1^4 +0.05\times q_2^4}{4}$ & 2\\
        Henon Heiles~\cite{henonheiles1964} & $H(q_1,q_2,p_1,p_2) = \frac{p_1^2 + p_2^2}{2} + \frac{q_1^2 + q_2^2}{2} + q_1^2 \times q_2 - \frac{q_2^3}{3}$ & 2\\
        Toda Lattice~\cite{Toda1967,Ford1973} & $H(q_1,q_2,q_3,p_1,p_2,p_3) = \frac{p_1^2+p_2^2+p_3^2}{2}+\exp(q_1-q_2)+\exp(q_2-q_3)+\exp(q_3-q_1)-3$ & 3\\
        Coupled Oscillator & $H(q_1,\dots,q_{n},p_1,\dots,p_{n}) = \sum^{n}_{i=1}\frac{p_i^2}{2} + \sum^{n}_{i=2}\frac{(q_i-q_{i-1})^2}{2}  $ & $n$\\
    \end{tabular}
\end{ruledtabular}
\end{table*}

For illustration and comparative empirical evaluation of the HNNs presented, we consider five separable Hamiltonian systems in Table~\ref{tab:equationsanddimensions}.
These allow the HNNs to demonstrate their performance in predicting a range of Hamiltonian dynamics comprising different functions, different values of $n$, and well-behaved or chaotic dynamics, from observations.

\section{Methodology}~\label{sec:methodology}

\subsection{The HNN Baseline} \label{sec:method:learningbias}

\begin{figure*}[ht]
    \centering
    \includegraphics[width = 0.89\textwidth]{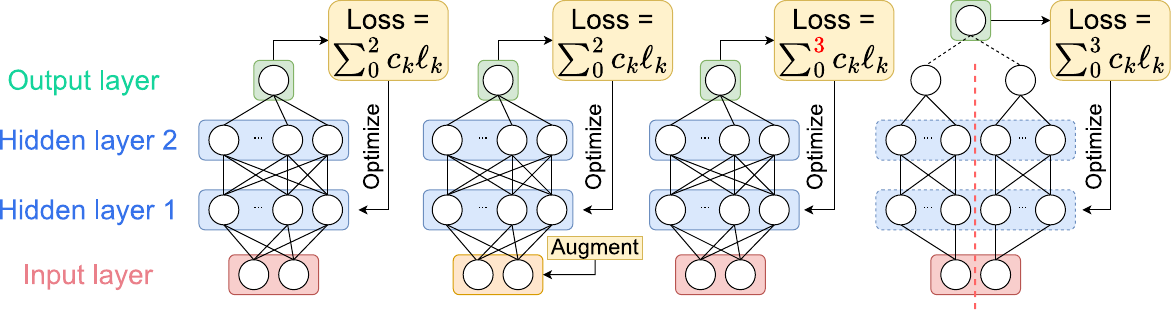}
    \caption{Leftmost: Architecture of the baseline HNN. Second from left: Proposed separable HNN with observational bias. Second from right: Proposed separable HNN with learning bias. Rightmost: Proposed separable HNN with inductive bias. }
    \label{fig:architecture}
\end{figure*}



The baseline HNN (leftmost, Fig.~\ref{fig:architecture}), proposed in ~\cite{bertalan_2019,Greydanus_hamiltoniannn_2019}, is described as follows,
\begin{eqnarray}\label{eqn:predictedsoln}
    \hat{H}(q,p;w) \approx H(q,p),
\end{eqnarray}
where $(q, p)$ is an input state. 
$\hat{H}$ is a neural network with trainable parameters $w$.
$\hat{H}(q,p;w)$ and $H(q,p)$ represent the HNN predicted Hamiltonian value and the true Hamiltonian value, respectively.

The predicted time derivatives of the HNN are
\begin{equation}
    \dot{\hat{q}} = \frac{\partial \hat{H}}{\partial p}, \quad 
    \dot{\hat{p}} = -\frac{\partial \hat{H}}{\partial q}, \label{eqn:predictedvector}
\end{equation}
which can be computed by automatic differentiation.

The loss function of the baseline HNN is
\begin{eqnarray}
    \ell(q, p, \dot{q}, \dot{p}; w) &= \sum^2_{k=0} c_k \ell_k, 
    \label{eqn:MLPloss1}
\end{eqnarray} 
where $c_k\in \mathbb{R}$ is the coefficient of each $\ell_k$, and
\begin{eqnarray}
    \ell_0(q, p, \dot{q}, \dot{p}; w) &=& \left( \hat{H}({q_0},{p_0}; w) - H_0 \right)^2, \label{eqn:ell0} \\
    \ell_1(q, p, \dot{q}, \dot{p}; w) &=& \left\| \dot{\hat{q}} - \dot{q} \right\|^2, \label{eqn:ell1}\\
    \ell_2(q, p, \dot{q}, \dot{p}; w) &=& \left\| \dot{\hat{p}} - \dot{p} \right\|^2. \label{eqn:ell2}
\end{eqnarray}
$\ell_0$ is an arbitrary pinning term that "pins" the regressed Hamiltonian at $(q_0,p_0)$ to one solution, $H_0\in \mathbb{R}$, among several solutions that are modulo and additive constant and reduces the search space for convergence. 
$\ell_1$ and $\ell_2$ are Hamilton's equations corresponding to Eq.~\eqref{eqn:defnHamiltonian2}. $\ell_0$, $\ell_1$, and $\ell_2$ introduce biases that favor the convergence of the HNN toward the underlying physics of the regressed Hamiltonian. 
Eq.~\eqref{eqn:MLPloss1} defines the loss function of the HNN as a linear combination of Eq.~\eqref{eqn:ell0} to Eq.~\eqref{eqn:ell2}. 
One can assume that $c_k = 1$ although additional knowledge of the system can be used to emphasize any $\ell_k$~\cite{bertalan_2019}.

The separable HNNs introduced in Sec.~\ref{sec:method:addsep_obsbias}, \ref{sec:method:addsep_learningbias}, \ref{sec:addsep_inductivebias} and \ref{sec:method:mixedbias} adopt the Hamiltonian learning bias of the HNN and further embed biases regarding additive separability. They perform the task of regressing the Hamiltonian and vector field in the same way.

\subsection{The HNN with Observational Bias (HNN-O)} 
\label{sec:method:addsep_obsbias}
We inform the baseline HNN of additive separability via data augmentation. 
Given a dataset of instantaneous states $(q,p)$ and time derivatives $(\dot{q},\dot{p})$, 
we augment the dataset and create new samples. 
The derivatives of the new samples $(\dot{q},\dot{p})$ are dependent solely on $p$ or $q$ respectively, and can be inferred from the original samples.

\begin{figure}[h]
\centering
    \includegraphics[width =0.2\textwidth]{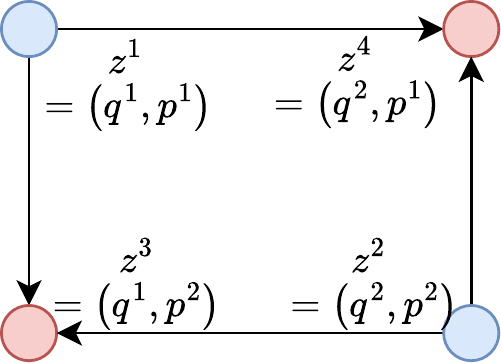} 
    \hspace{4mm}
    \includegraphics[width =0.2\textwidth]{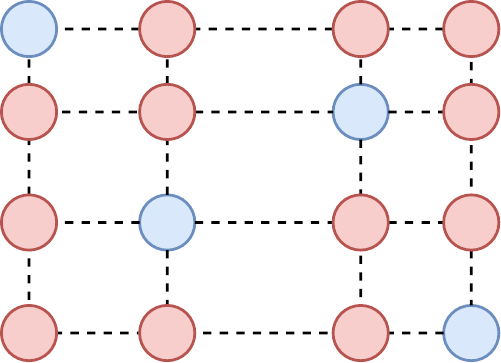} 
    \label{fig:observationalbias}
       \caption{\label{fig:OB1}Left: 2 new samples (in red) are created from 2 old samples (in blue). \label{fig:OB2}Right: 12 new samples are created from 4 samples. Generally, quadratically more new samples are created. }
\end{figure}

Consider two samples comprising their states and derivatives denoted by: $(z^1=(q^1, p^1), \dot{z}^1=(\dot{q}^1, \dot{p}^1))$ and $(z^2=(q^2, p^2), \dot{z}^2=(\dot{q}^2, \dot{p}^2))$.
The superscript represents different samples in the dataset. Based on these two, the observational bias creates two new samples, denoted by $(z^3, \dot{z}^3)$ and $(z^4, \dot{z}^4)$.
The states $z^3$ and $z^4$ are:
\begin{equation}\label{eq: z3z4}
    z^3 = (q^1, p^2),\quad z^4 = (q^2, p^1),
\end{equation}
respectively. 
By Eq. \eqref{eqn:defnHamiltonian2}, the derivatives $\dot{z}^3$ and $\dot{z}^4$ are:
\begin{equation}\label{eq: z3z4_dot}
\begin{aligned}
    \dot{z}^3 = \left(\frac{\partial H}{\partial p}(q^1, p^2), -\frac{\partial H}{\partial q}(q^1, p^2) \right), \\
    \dot{z}^4 = \left(\frac{\partial H}{\partial p}(q^2, p^1), -\frac{\partial H}{\partial q}(q^2, p^1) \right),
\end{aligned}
\end{equation}
respectively. 

Using the fact that the Hamiltonian system is additive separable, we have 
\begin{equation}
    \begin{aligned}
    \frac{\partial H}{\partial p}(q^1, p^2) 
     &=\frac{\partial H}{\partial p} (q^2, p^2) &=\dot{q}^2,\\
    -\frac{\partial H}{\partial q}(q^1, p^2) &= -\frac{\partial H}{\partial q} (q^1, p^1) &=\dot{p}^1,\\
    \frac{\partial H}{\partial p}(q^2, p^1) &= \frac{\partial H}{\partial p} (q^1, p^1) &=\dot{q}^1,\\
    -\frac{\partial H}{\partial q}(q^2, p^1) &= -\frac{\partial H}{\partial q} (q^2, p^2) &=\dot{p}^2.
\end{aligned}
\end{equation}

Then, Eq. \eqref{eq: z3z4_dot} becomes 
\begin{eqnarray}
    \dot{z}^3 = \left(\dot{q}^2, \dot{p}^1\right),
    \quad
    \dot{z}^4 = \left(\dot{q}^1, \dot{p}^2 \right).
\end{eqnarray}
Therefore, the quantity of data has been doubled from two samples to four samples, as shown in Fig.~\ref{fig:OB1} (left).


Generally, given data $\mathcal{M} = \{(q^i, p^i, \dot{q}^i, \dot{p}^i)\}_{i=1}^{K}$, the quantity of data can be increased $\mu$ times, where $\mu = 1, 2, \dots, K-1$.
For each $\mu$, a new data set $\{(q^i,p^{i+\mu}, \dot{q}^{i+\mu}, \dot{p}^i)\}_{i=1}^K$ is created.
We then append all $\mu$ sets of new data to the original data. At most $K\times (K-1)$ new data can be created from the original data, as shown in Fig.~\ref{fig:OB2} (right).



The observational bias creates more data to improve coverage of the input domain of the Hamiltonian regression task, and the regressed HNN reflects the additive separability of the data. This improves regression performance but increases the time taken for forward and backward propagation of the separable HNN. In selecting the optimal number of samples, there is a trade-off between regression performance and training time. 

\subsection{The HNN with Learning Bias (HNN-L)} 
\label{sec:method:addsep_learningbias}
We inform the baseline HNN of additive separability via its loss function, as given below:

\begin{equation}
    \ell_3(q, p; w) =  \left\|\frac{\partial^2 \hat{H}}{\partial q \partial p} (q, p; w) \right\|^2 \label{eqn:MLPloss2}
\end{equation}
\begin{eqnarray}
    \ell_{sep}(q,p,\dot{q},\dot{p}; w) = \sum^3_{k=0} c_k \ell_k. \label{eqn:seploss}
\end{eqnarray}

$\ell_3$ in Eq.~\eqref{eqn:MLPloss2} is the Frobenius matrix norm of the mixed partial derivative of $\hat{H}$, corresponding to Eq.~\eqref{eqn:addsep_hamiltonian_mixedpartial}. $\ell_{sep}$ is the loss function of the separable HNN with learning bias. It is a linear combination of equations $\ell_0$ to $\ell_3$ from Eqs.~\eqref{eqn:MLPloss1} and \eqref{eqn:MLPloss2}. 
The resulting HNN-L with learning bias (second from right, Fig.~\ref{fig:architecture}) favors convergence towards a HNN that is additively separable. 
A larger $c_3$ allows the separable HNN to emphasize $\ell_3$ and additive separability of the HNN, but may also decrease the emphasis of $\ell_0$, $\ell_1$ and $\ell_2$, presenting a trade-off in the optimal value of $c_3$. 

\subsection{The HNN with Inductive Bias (HNN-I)} \label{sec:addsep_inductivebias}
We inform the baseline HNN of additive separability via its architecture. 
The proposed HNN-I (rightmost, Fig.~\ref{fig:architecture}) is described as follows,
\begin{eqnarray}
    \hat{H}_{IB}(q,p;w) = \hat{T}(q;w_1) + \hat{V}(p;w_2) \approx H(q,p), \label{eqn:HNNI}
\end{eqnarray}
where $\hat{T}$ and $\hat{V}$ are two independent neural networks, conjoined only at the output layer, with trainable parameters $w_1$ and $w_2$, respectively.
$w$ is composed of $w_1$ and $w_2$. The mixed partial derivative of the HNN-I is always zero and additive separability is strictly satisfied. 






The conjoined neural networks can be trained consecutively in series, or simultaneously in parallel. Training in series refers to propagating through $\hat{T}$ and $\hat{V}$ separately. The HNN-I trained in series requires twice as many forward propagations as the HNN.
Training in parallel refers to propagating through $\hat{T}$ and $\hat{V}$ at the same time. It halves the number of forward propagations through the conjoined neural networks as compared to training in series. However, each forward propagation computes matrix multiplications with higher computational complexity as the weight matrix is higher dimensional. This presents a trade-off in finding the optimal implementation to minimize training time. 

One may also vary the summation layer of the separable HNN, to utilize a simple sum or introduce weights and biases. The different implementations for the summation layer present a trade-off in finding the optimal implementation of the HNN-I. The impact of this decision is relatively minor; hence, we have included the detailed experiments and discussion in Appendex~\ref{sec:lastlayer}.

\subsection{The HNN with Multiple Biases} \label{sec:method:mixedbias}
As seen from Sec.~\ref{sec:method:addsep_obsbias}, \ref{sec:method:addsep_learningbias} and \ref{sec:addsep_inductivebias}, multiple biases can be individually embedded into a HNN. The biases can be further combined to constrain a system to be additively separable. Multiple biases regarding the additive separability of a Hamiltonian to be regressed can be embedded into a neural network simultaneously.

Pairwise combinations of the observational, learning, and inductive biases are first considered. There are three pairwise combinations of biases. The HNN-OL combines observational and learning biases, resulting in a separable HNN with increased training data and a loss function given by Eq.~\eqref{eqn:seploss}. The HNN-LI combines the learning and inductive biases, resulting in a separable HNN with a loss function given by Eq.~\eqref{eqn:seploss} and an architecture comprising two smaller neural networks, conjoined at the output layer. The third HNN-OI combines observational and inductive biases, resulting in a separable HNN with increased training data and an architecture comprising two smaller neural networks, conjoined at the output layer. 

Lastly, the HNN-OLI combines observational, learning, and inductive biases, resulting in a separable HNN with increased training data, a loss function given by Eq.~\eqref{eqn:seploss} and an architecture comprising two smaller neural networks, conjoined at the output layer. 

\section{Experiments for General Systems} ~\label{sec:exp}
The task is to regress the Hamiltonian following Eq.~\eqref{eqn:predictedsoln} and the vector field following Eq.~\eqref{eqn:predictedvector}. We compare the performance of the baseline HNN and the proposed separable HNNs on the task, for the five additively separable Hamiltonian systems shown in Table~\ref{tab:equationsanddimensions}. 
In particular, the Coupled Oscillator system considered in this section is $n=3$.



We use the following two evaluation metrics. 
\begin{itemize}
    \item 
    For the regression of the Hamiltonian: 
    \begin{eqnarray}
        E_H =\frac{1}{K} \sum^{K}_{k=1} \left(\hat{H}(q^k, p^k; w)-H(q^k, p^k)\right)^2,  \label{eqn:abserror}
    \end{eqnarray}
    where $\hat{H}(q^k, p^k; w)$ is the predicted  Hamiltonian by the HNN, $H(q^k, p^k)$ is the true Hamiltonian, and $K$ is the size of a test batch.
    \item For the regression of the vector field:
\begin{eqnarray}
    E_V =
    \frac{1}{K} \sum^{K}_{k=1} \frac{\sqrt{||\dot{\hat{q}}^k-\dot{q}^{k}||_2^2+||\dot{\hat{p}}^k-\dot{p}^{k}||_2^2}}{\sqrt{||\dot{q}^{k}||_2^2+||\dot{p}^{k}||_2^2}} \times 100\%,
    \label{eqn:vectorerror}
\end{eqnarray}
where $\dot{\hat{q}}^k$ and $\dot{\hat{p}}^k$ are the predicted time derivatives, and $\dot{q}^k$ and $\dot{p}^k$ are the corresponding true derivatives.
$K$ is the size of a test batch.
\end{itemize}

\begin{table*}
\caption{\label{tab:nnsetup}Configuration of experiments. $\Omega_q$ and $\Omega_p$ are the domains of $q$ and $p$ respectively. The last four columns show the width and total parameters for the various HNNs for each experiment. For all experiments, by design, the number of parameters for the HNN-I is slightly less than the number of parameters for the baseline HNN, HNN-O, and HNN-L.}
\begin{ruledtabular}
\begin{tabular}{cccccccc}
        \multicolumn{1}{c}{\multirow{2}{*}{Experiment}}   & \multicolumn{1}{c}{\multirow{2}{*}{$\Omega_q$}} & \multicolumn{1}{c}{\multirow{2}{*}{$\Omega_p$}} & \multicolumn{1}{c}{\multirow{2}{*}{Learning rate}}  & \multicolumn{2}{c}{HNN, HNN-O,HNN-L} & \multicolumn{2}{c}{HNN-I}  \\
        \multicolumn{1}{c}{}                              & \multicolumn{1}{c}{}                            & \multicolumn{1}{c}{}                            & \multicolumn{1}{c}{} & Width   & Params                     & Width & Params \\
        \hline
        Nonlinear Pendulum & $[-2\pi, 2\pi]$ & $[-1.2, 1.2]$ & 0.001 & 16~\cite{bertalan_2019} & 337 & 11 & 332 \\
        Anisotropic Oscillator & $[-0.5, 0.5]$ & $[-0.5, 0.5]$ & 0.001 & 32 & 1249 & 22 & 1190 \\
        Henon Heiles & $[-0.5, 0.5]$ & $[-0.5, 0.5]$ & 0.01 & 32 & 1249 & 22 & 1190 \\
        Toda Lattice & $[-0.5, 0.5]$ & $[-0.5, 0.5]$ & 0.01 & 31 & 1241 & 22 & 1234\\
        Coupled Oscillator & $[-0.5, 0.5]$ & $[-0.5, 0.5]$ & 0.01 & 31 & 1241 & 22 & 1234\\
    \end{tabular}
\end{ruledtabular}
\end{table*}

The data used for the experiments is as follows. Training data comprising $512$ samples of $(q,p,\dot{q},\dot{p})$ are generated uniformly at random within the sampling domain for each Hamiltonian system. The sampling domains are shown in Table~\ref{tab:nnsetup}. 20\% of training data is set aside as validation data for validation-based early stopping~\cite{Prechelt1998}. The test data set comprises $10^{2n}$ evenly spaced samples of $(q,p,\dot{q},\dot{p})$. The training and testing data is available at \verb|github.com/zykhoo/SeparableNNs|.

The general experimental setup for all HNNs in all experiments is as follows. 
All HNNs are designed with an input layer with width $2\times n$, two hidden layers with width shown in Table~\ref{tab:nnsetup}, and one output layer with width one. They use a softplus activation. For training, a batch size of 80 is used, with an Adam optimizer~\cite{kingma2014adam}, with a learning rate shown in Table~\ref{tab:nnsetup}. All experiments are repeated for 20 random seeds. The random seeds are used for the weight initialization of the HNNs and the sampling of training data. 
All HNNs are trained in \verb|Pytorch|. 
We perform automatic differentiation using tools such as \verb|Pytorch|~\cite{pytorch-autodiff} or \verb|JAX|~\cite{jax2018github} to compute the $\dot{\hat{q}}$ and $\dot{\hat{p}}$. 
The complete code in Python is available at \url{github.com/zykhoo/SeparableNNs}. 

In all experiments, unless stated otherwise, all HNNs are trained until convergence. In our context, convergence is defined when the validation loss does not decrease for 4,000 epochs. This is to find an optimal bias-variance trade-off. 

The experiments in this Section comprise three parts. Sec.~\ref{sec:hyperparam} tunes the hyperparameters of the proposed separable HNNs. 
Sec.~\ref{sec:expregression} and Sec.~\ref{sec:exptimebudget} compare the proposed separable HNNs against the baseline HNN in terms of accuracy and efficiency, respectively.

\subsection{Hyperparameter Tuning} \label{sec:hyperparam}


\begin{figure*}[t]
    \centering
    \includegraphics[width = 1.0\textwidth]{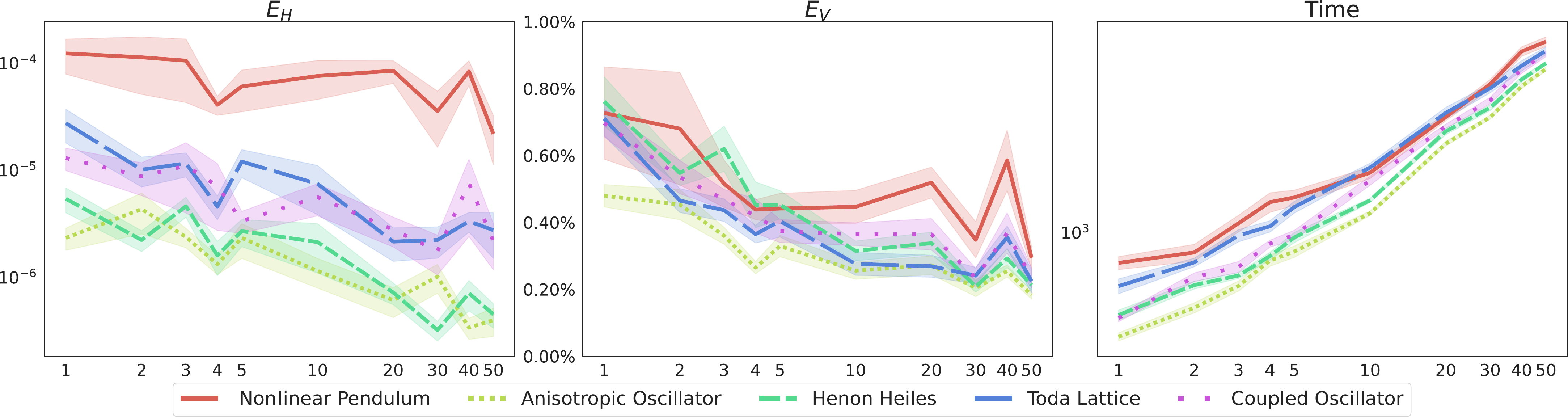}
    \caption{
    Performance of the HNN-O with $\mu\in \{1,2,3,4,5,10,20,30,40,50\}$ for the Nonlinear Pendulum (red, solid line), Anisotropic Oscillator (yellow, densely dotted line), Henon Heiles (green, loosely dashed line), Toda Lattice (blue, densely dashed line), and Coupled Oscillator (purple, loosely dotted line) systems. Left: $E_H$ and standard error. Center: $E_V$ and standard error. Right: Wall-clock time taken until convergence in seconds and standard error.}
\label{fig:observational_error}
\end{figure*}

We focus on the HNN-O proposed in Sec. \ref{sec:method:addsep_obsbias}.
We aim to determine the optimal number of new samples to create with training data $\mathcal{M}$.
Excluding the validation data, $K=80\% \times 512= 410$ training samples can be used to create new training samples. $\mu$ is the number of times that the training data can be increased. We consider $\mu\in \{1,2,3,4,5,10,20,30,40,50\}$ to augment the training data. The separable HNN with observational bias is set up following the leftmost architecture in Fig.~\ref{fig:architecture}. 
Fig.~\ref{fig:observational_error} reports $E_H$, $E_V$, and training time for the HNN-O. 
Generally, as $\mu$ increases, the $E_H$ and $E_V$ decrease, but the training time increases. The marginal improvement in $E_H$ and $E_V$ decreases as $\mu$ increases. The training time generally increases proportionally to $\mu$ because the time taken for each training epoch increases proportionately to $\mu$. The convergence time of the HNN-O with $\mu=1$ is approximately one order of magnitude faster than the HNN-O with $\mu=50$. A good trade-off that balances reducing the vector error and training time is the HNN-O with $\mu=2$. 






\begin{figure*}
    \centering
    \includegraphics[width = \textwidth]{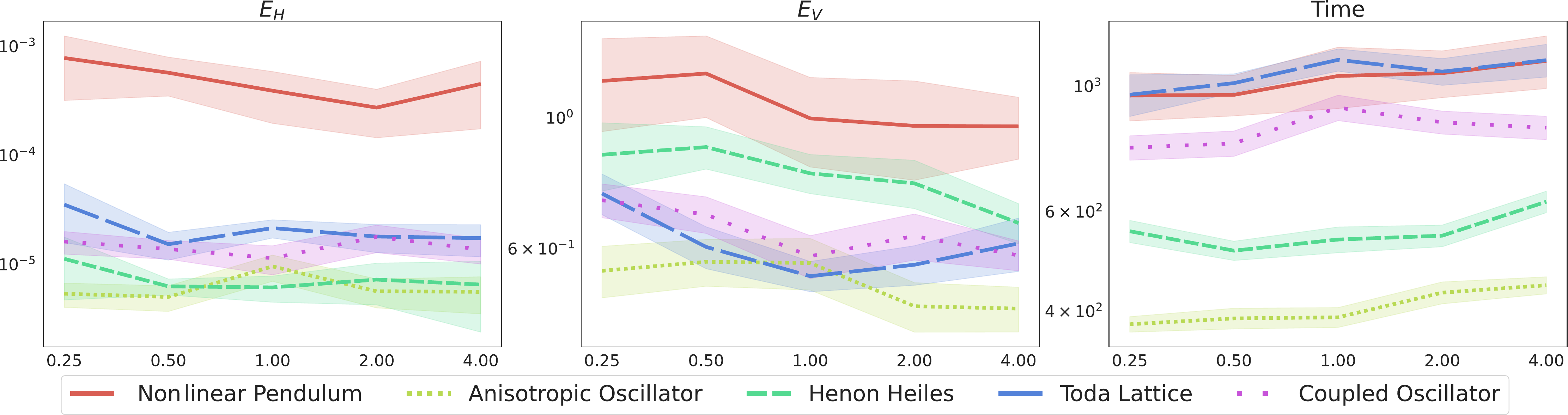}
    \caption{
    Performance of the HNN-L with $c_3\in \{0.25, 0.50, 1.00, 2.00, 4.00\}$ for the Nonlinear Pendulum (red, solid line), Anisotropic Oscillator (yellow, densely dotted line), Henon Heiles (green, loosely dashed line), Toda Lattice (blue, densely dashed line), and Coupled Oscillator (purple, loosely dotted line) systems. Left: $E_H$ and standard error. Center: $E_V$ and standard error. Right: Wall-clock time taken until convergence in seconds and standard error.}
    \label{fig:learning_error}
\end{figure*}
We focus on the HNN-L proposed in Sec. \ref{sec:method:addsep_learningbias}.
We study the trade-off between emphasizing additive separability in $\ell_3$ and the Hamiltonian properties in $\ell_0$, $\ell_1$ and $\ell_2$ in Eq.~\eqref{eqn:MLPloss2}. 
The HNN-L is trained and evaluated for cases where the weighting parameter $c_3 \in \{0.25, 0.50, 1.00, 2.00, 4.00\}$, and the optimal value of $c_3$ is determined empirically.
Fig.~\ref{fig:learning_error} reports the $E_H$, $E_V$, and the training time for the HNN-L.
Generally, from $c_3=0.25$ and $c_3=4.00$, the $E_V$ decreases with increasing $c_3$ while the training time required increases. The performance of the HNN-L with $c_3=0.50$, $c_3=1.00$, and $c_3=2.00$ do not strictly follow this trend, but these suggest that as the value of $c_3$ increases, the emphasis on additive separability increases, and the HNN-L places less emphasis on learning Hamilton's equations. As a result, the number of epochs required for the HNN-L to converge increases, increasing training time. 
The HNN-L with $c_3 = 4.00$ generally regresses the vector field best. The HNN-L with $c_3 = 0.25$ is generally the fastest. A good trade-off that balances the importance of additive separability and Hamilton's equations appears to be $c_3=1.00$. 

We focus on the HNN-I proposed in Sec. \ref{sec:addsep_inductivebias}.
We aim to determine whether the HNN-I should be trained in series or parallel.
The HNN-I is set up following the rightmost architecture in Fig.~\ref{fig:architecture}. The width of each smaller conjoined neural network is shown in Table~\ref{tab:nnsetup}. 
The proposed HNN-I can be trained consecutively. In this case, the HNN-I is composed of two independent neural networks. Forward propagation is through each conjoined neural network separately. Propagation occurs through $\hat{T}(q)$ first, then $\hat{V}(p)$. Their sum is the predicted Hamiltonian. Alternatively, the proposed HNN-I can be trained simultaneously and in parallel. Each hidden layer of the HNN-I trained in parallel is a stack of two layers, one layer from $\hat{T}(q)$ and one layer from $\hat{V}(p)$. Forward propagation occurs through these stacked layers. As compared to the HNN-I trained in series, the HNN-I trained in parallel operates like one wider neural network that computes $\hat{T}(q)$ and $\hat{V}(p)$, stacked. 
\begin{figure*}
    \centering
    \includegraphics[width = \textwidth]{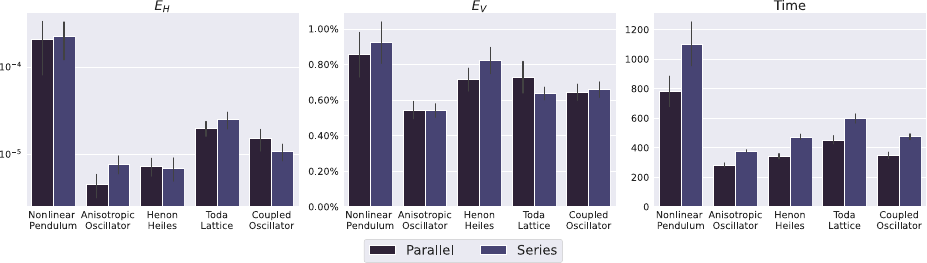}
    \caption{Performance of the HNN-I trained in parallel or in series. Left: $E_H$ and standard error. Center: $E_V$ and standard error. Right: Wall-clock time taken until convergence in seconds and standard error. The bar plots in each group are presented in the order shown in the legend.}
\label{fig:inductive_error_body}
\end{figure*}
Fig.~\ref{fig:inductive_error_body} reports the $E_H$, $E_V$, and time taken for the body of the HNN-I. It is observed that the HNN-I trained in parallel is consistently faster than that which is trained consecutively. 
Furthermore, it is observed that training the conjoined neural networks in parallel and in series results in different $E_H$ and $E_V$. This is counter-intuitive but closer analysis reveals this may be due to different floating point errors which cause their $E_H$ and $E_V$ to diverge after many iterations. Generally, both HNN-I perform well and have similar $E_V$. As a result, the HNN-I trained in parallel is preferred.



\subsection{Accuracy}\label{sec:expregression}
We focus on comparing the HNNs on the task of regressing the Hamiltonian and vector field. Eight HNNs are compared. They are the baseline HNN, the three separable HNNs with one bias each namely the HNN-O, HNN-L, and HNN-I, the three separable HNNs with two biases each namely the HNN-OL, HNN-LI, and HNN-OI, and the separable HNN with all three biases, the HNN-OLI. 
From Sec. \ref{sec:hyperparam}, the optimal implementations of the various separable HNNs are used. All details of the setup, training, and evaluation follow from Sec.~\ref{sec:exp}.  

\begin{figure*}
    \centering
    \includegraphics[width = \textwidth]{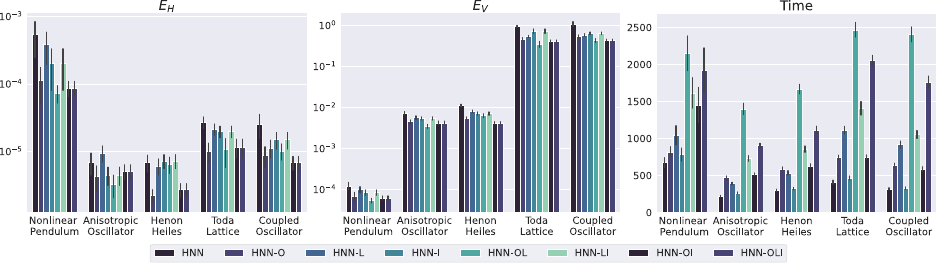}
    \caption{Performance of the eight HNNs. Left: $E_H$ and standard error. Center: $E_V$ and standard error. Right: Wall-clock time taken until convergence in seconds and standard error. The bar plots in each group are presented in the order shown in the legend.}
    \label{fig:all_error}
\end{figure*}


Fig.~\ref{fig:all_error} reports $E_H$, $E_V$, and the time taken to train each HNN. 
We observe that all separable HNNs have a lower $E_H$ and $E_V$ than the HNN. Among the three separable HNNs with one bias, the HNN-O has the lowest $E_H$ and $E_V$. Among the three HNNs with two biases, either the HNN-OL or HNN-OI has the lowest $E_H$ and $E_V$. 
The proposed HNNs leverage physics information regarding separability to penalize or prevent interaction between the states and this reduces the complexity of the Hamiltonian regression problem. As a result, both $E_H$ and $E_V$ are reduced. 

The HNN-O performs the best among the separable HNNs with one bias as it generates more samples of the data. This emphasizes additive separability and covers the input domain of the Hamiltonian system. This eases the interpolation task of the neural network. 
The HNN-I generally outperforms the HNN-L as it restricts regressions to strictly satisfy separability, forcing the HNN-I to simplify a complex regression problem into two smaller ones. 
We also observe that among the three proposed HNNs with one bias, the HNN-I is the fastest. Among the three proposed separable HNNs with two biases, the HNN-OI is the fastest. 
The HNN-I is slower than the baseline HNN. The HNN-I has a conjoined architecture with higher dimensional weight and bias matrices. These slightly increase each forward and backward propagation time. The HNN-L is even slower as it computes Eq.~\eqref{eqn:MLPloss2}. The HNN-O is the slowest as it has several times more samples that linearly scale the training time per epoch given the same batch size. Among the three separable HNNs, the HNN-I has the fastest time per epoch.

We make two observations about the inductive bias. Firstly, the HNN-I and HNN-LI have the same $E_H$ and $E_V$. Likewise, the HNN-OI and HNN-OLI also have the same $E_H$ and $E_V$. The learning bias becomes redundant when paired with an inductive bias. However, the HNNs with learning biases still spend computational resources computing Eq.~\eqref{eqn:MLPloss2}. Secondly, a useful property of separable HNNs with inductive biases is that it allows the recovery and interpretation of the kinetic and potential energies of a Hamiltonian system. This property is explored in detail in Appendix~\ref{sec:expKEPE}. 

In conclusion, the HNN-OI is the optimal HNN and our HNN of choice. It consistently outperforms the baseline HNN and is relatively fast compared to the other proposed separable HNNs. However, it is unclear if the HNN-OI, or other separable HNNs can outperform the baseline HNN if both HNNs are limited to the same computational resources. This motivates the experiments in Sec.~\ref{sec:exptimebudget}, where the HNNs are compared when trained with the same computational resources.

\subsection{Efficiency}~\label{sec:exptimebudget}
We focus on comparing the HNNs on the task of regressing the Hamiltonian and vector field under a time budget. 
The experimental setup is modified from Sec.~\ref{sec:expregression} to account for a time budget. The time budget is a measure of the computational resources each HNN uses for training. The time budget is set as the time taken for the baseline HNN to converge. 
Each separable HNN ends its training once the time budget is reached.

\begin{figure*}
    \centering
    \includegraphics[width = \textwidth]{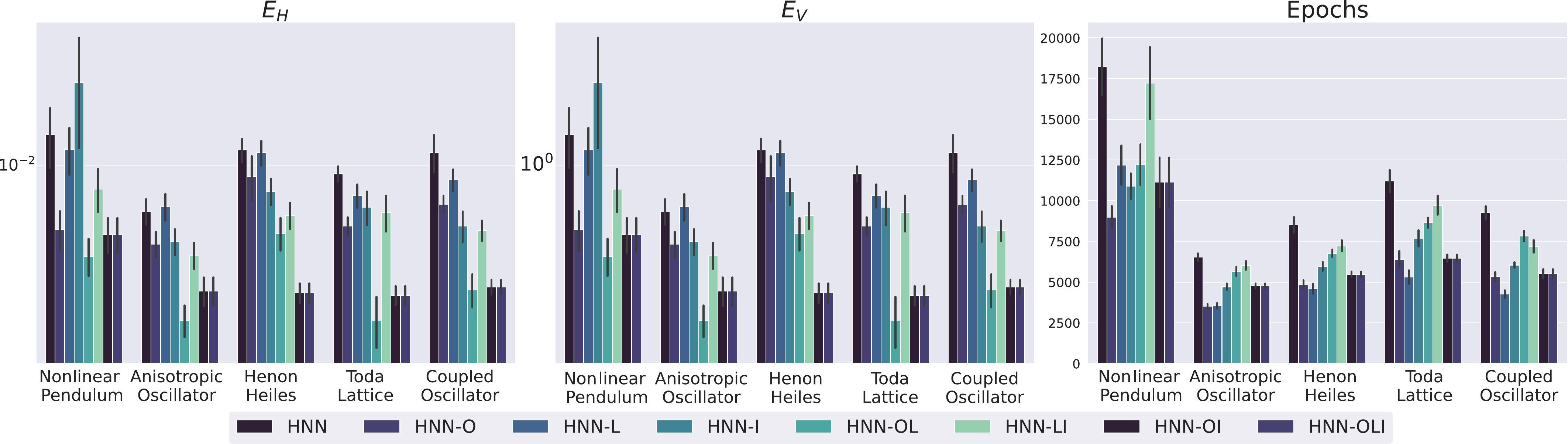}
    \caption{Performance of the eight HNNs evaluated with a time budget. Left: $E_H$ and standard error. Center: $E_V$ and standard error. Right: Number of epochs completed and standard error. The bar plots in each group are presented in the order shown in the legend.}
    \label{fig:timebudget_vectorerror}
\end{figure*}


Fig.~\ref{fig:timebudget_vectorerror} reports the $E_H$, $E_V$, and number of epochs for each HNN trained within the time budget for all HNNs.
Generally, among HNNs with one bias, the HNN-O consistently outperforms the HNN. 
Among the HNNs with two biases, the HNN-OI and HNN-OL generally perform the best and consistently outperform the HNN. They are the best-performing among all HNNs. As the HNN-OI is the best-performing in Sec.~\ref{sec:expregression}, this shows that it requires significantly more computational resources to converge. 
Lastly, we note that within the time budget given by the baseline HNN, most separable HNNs complete fewer epochs. The separability biases increase the time or computational resources required for each epoch. 
Nonetheless, the separable HNNs still have a lower $E_H$ and $E_V$ than the baseline HNN within the time budget. This is because the separability bias allows the training loss to fall faster with each epoch.

In summary, the proposed separable HNNs leverage physics information regarding separability to improve the regression of the Hamiltonian and vector field. The observational bias allows the acquisition of training data to cover the input domain of the Hamiltonian system and reinforce the additive separability bias. The learning bias places a soft constraint of additive separability on the learned Hamiltonian and vector field. The inductive bias restricts regressions of the Hamiltonian and vector field to strictly satisfy separability. These biases allow the separable HNNs to converge to lower $E_H$ and $E_V$. The penalty for these biases is an increase in training time. Nonetheless, combining these biases yields the best HNN, in the form of the HNN-OI.


We further focus on comparing the HNN-OI and baseline HNN on their convergence during training. 
The experimental setup is modified from Sec.~\ref{sec:expregression}. We redefine convergence in this experiment to be when the model has trained for 15,000 epochs. The model is saved every 50 epochs during training, and the model is evaluated on the test data set to visualize the reduction in $E_H$ and $E_V$ with epochs. 

\begin{figure*}[hbt!]
    \centering
    \includegraphics[width = \textwidth]{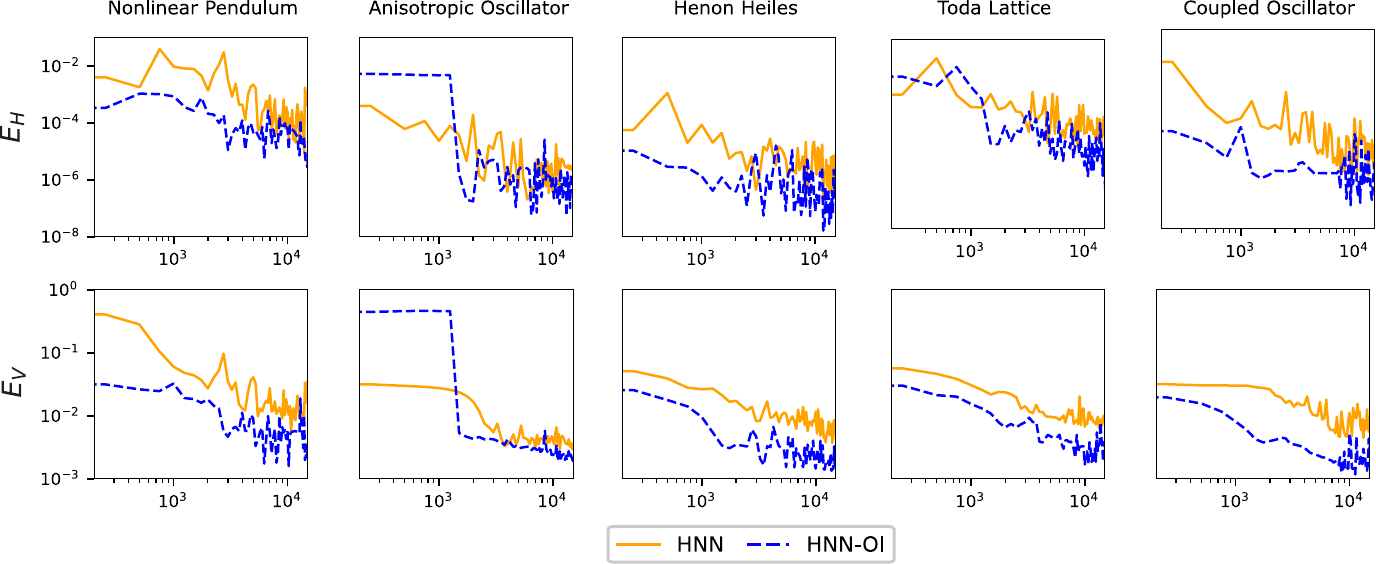}
    \caption{Convergence of the HNN-OI (blue, dashed line) and HNN (orange, solid line) against the number of epochs. Top: The y-axis represents $E_H$. The x-axis represents the number of epochs trained by the model. Bottom: The y-axis represents $E_V$. The x-axis represents the number of epochs trained by the model.}
    \label{fig:convergence}
\end{figure*}

Fig.~\ref{fig:convergence} plots the $E_H$ and $E_V$ against the number of epochs for the baseline HNN and the best-performing proposed separable HNN, the HNN-OI, for one random seed. Generally, the HNN-OI converges faster than the HNN. Furthermore, the steepest descent of the HNN-OI typically occurs within the first 2,000 epochs. When convergence is defined as the validation loss not decreasing for 4,000 epochs, all models would have experienced their steepest descent. Therefore, 4,000 epochs would be sufficient for all models to converge, and the HNN-OI outperforms the HNN. 
Generally, $E_H$ for the HNN-OI is also lower than the HNN. However, the descent is not as steep because $E_H$ is not explicitly minimized in the loss function.

\section{Experiments for Challenging Systems}~\label{sec:timedynamics}
\subsection{Chaotic Dynamics}
We focus on the chaotic Henon Heiles system to empirically demonstrate the accuracy of the proposed HNN-OI by comparing its dynamics against the baseline HNN. The vector fields from the HNNs trained in Sec.~\ref{sec:exp} for one random seed are numerically integrated from an initial condition using an Euler symplectic integrator~\cite{LEIMKUHLER1994117} to obtain their dynamics.   
The initial conditions are $(q_1, q_2, p_1, p_2) = (0.6, -0.3, 0.2, 0.2)$ corresponding to the energy or Hamiltonian value of $0.166$. The maximal Lyapunov exponent for this set of initial conditions is $0.139$ and since it is positive, the dynamics are chaotic~\cite{Shevchenko2003}. The Euler symplectic integrator numerically integrates this trajectory for the time interval $t=[0,6\pi]$ with a time step of $0.01$. 
\begin{figure}
    \centering
    \includegraphics[width = \columnwidth]{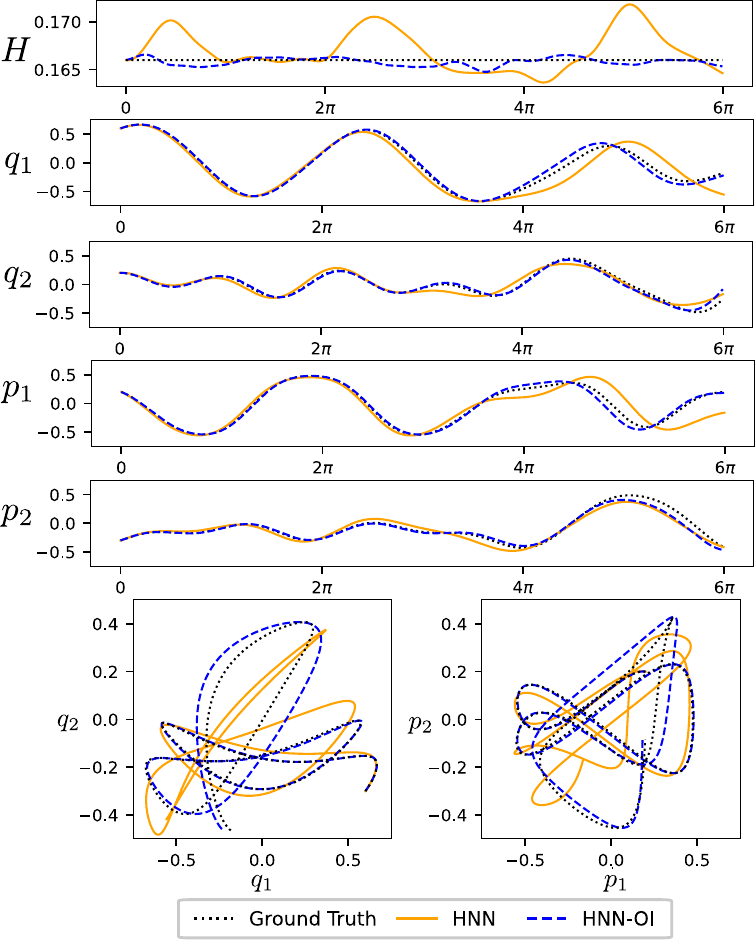}
    \caption{The trajectory of the Henon Heiles system was obtained by integrating the ground truth equations of motion (in black, dotted line), the vector field from the HNN-OI (orange, solid line), and the vector field from the HNN (blue, dashed line). Top: Dynamics over time. In each plot, the y-axis represents the value of $H$, $q$ or $p$. The x-axis represents the integration time. Bottom: The dynamics in selected planes. Bottom left: The $q_1$-$q_2$ plane. Bottom right: The $p_1$-$p_2$ plane.}
    \label{fig:HH_time}
\end{figure}

Fig.~\ref{fig:HH_time} reports the dynamics of the Henon Heiles system. We observe that generally, the HNN-OI predicts the dynamics of the Henon Heiles system better than the baseline HNN. The cumulative root mean squared error between the position of the HNN and the ground truth at every time step is 0.176. Between the HNN-OI and the ground truth, it is only 0.0698. The difference between the trajectory of the HNN and HNN-OI from the ground truth increases with time. This is because the error in predicting the trajectory accumulates with every integration step. The root mean squared error between the final position of the HNN and the ground truth is 0.531. The root mean squared error between the final position of the HNN-OI and the ground truth is 0.186. Lastly, we observe that the HNN-OI conserves the Hamiltonian better than the HNN. The cumulative root mean squared error between the Hamiltonian given by the HNN and the ground truth Hamiltonian at every time step is 0.0994. Between the HNN-OI and the ground truth, it is only 0.0192. Therefore, the HNN-OI is significantly better, and the model of choice, for predicting the dynamics of the Henon Heiles system.

\subsection{High Dimensional Dynamics}
We focus on the high dimensional Coupled Oscillator system to empirically demonstrate the accuracy of the proposed HNN-OI by comparing its dynamics against the baseline HNN. The Coupled Oscillator system in this experiment is modified from $n=3$ to $n=10$. The HNN trained on the high-dimensional Coupled Oscillator system had a width of 44 and 2968 trainable parameters. The HNN-OI had a width of 32 and 2900 trainable parameters. All other details of the training follow from Sec.~\ref{sec:exp}. The vector fields of both HNNs are numerically integrated from an initial condition using an Euler symplectic integrator~\cite{LEIMKUHLER1994117} to obtain their dynamics. 
For the Coupled Oscillator system, the initial conditions are a 20-dimension vector $(q_1,\dots q_{10}) = (0.6,-0.3,0.2,0.2,0.3,0.1,-0.2,0.3,0.2,-0.2)$ and $(p_1, \dots, p_{10}) = (0.3,0.1,-0.2,-0.2,0.2,-0.3,0.2,0.2)$. The Euler symplectic integrator numerically integrates this trajectory for the time interval $t=[0,6\pi]$ with a time step of $0.01$. 

\begin{figure}
    \centering
    \includegraphics[width = \columnwidth]{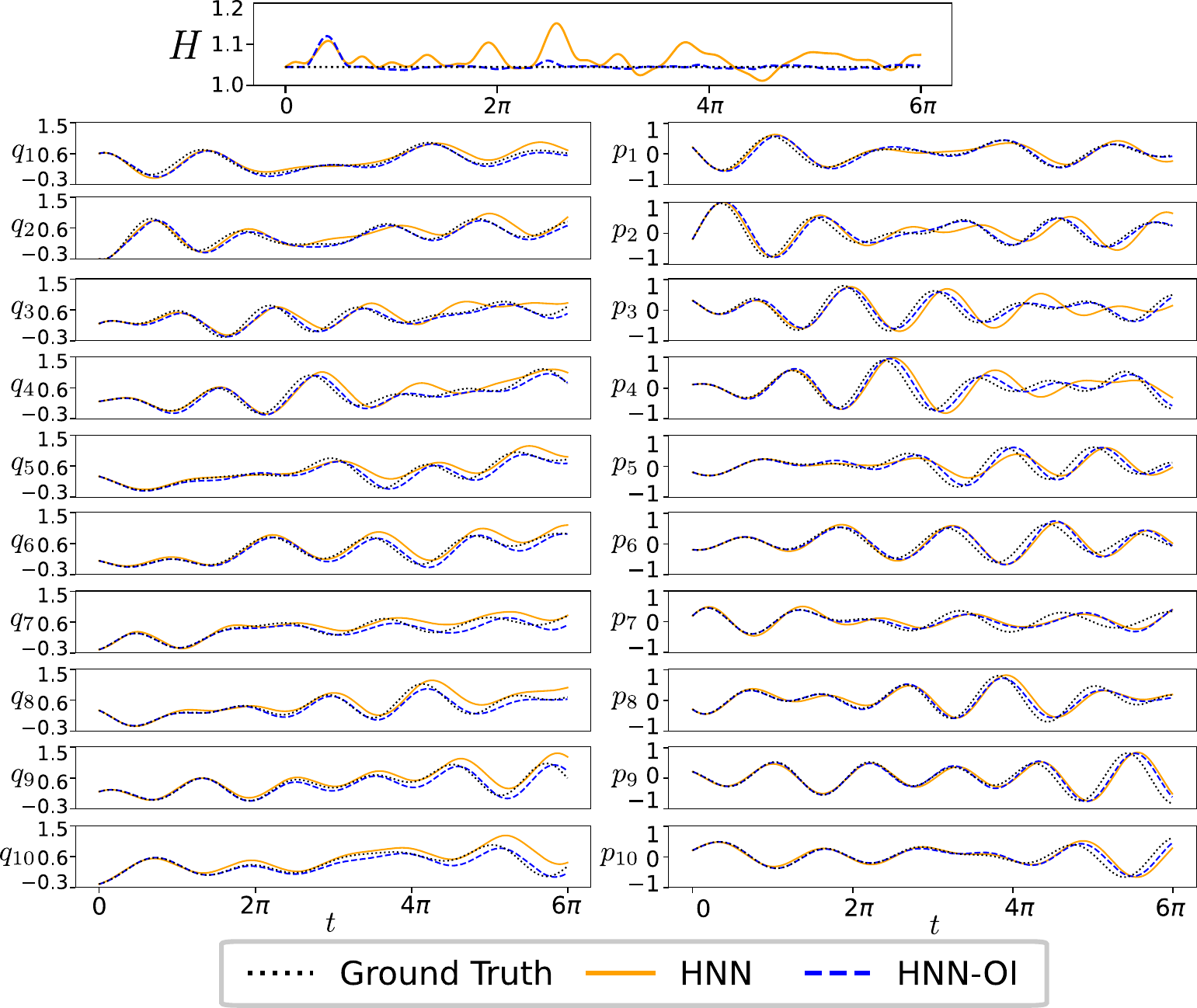}
    \caption{The trajectory of the Coupled Oscillator system with $n=10$ was obtained by integrating the ground truth equations of motion (in black, dotted line), the vector field from the HNN-OI (blue, dashed line), and the vector field from the HNN (orange, solid line). In each plot, the y-axis represents the value of $H$, $q$ or $p$. The x-axis represents the integration time.    }
    
    \label{fig:CO_time}
\end{figure}

\begin{figure*}
    \centering
    \includegraphics[width = \textwidth]{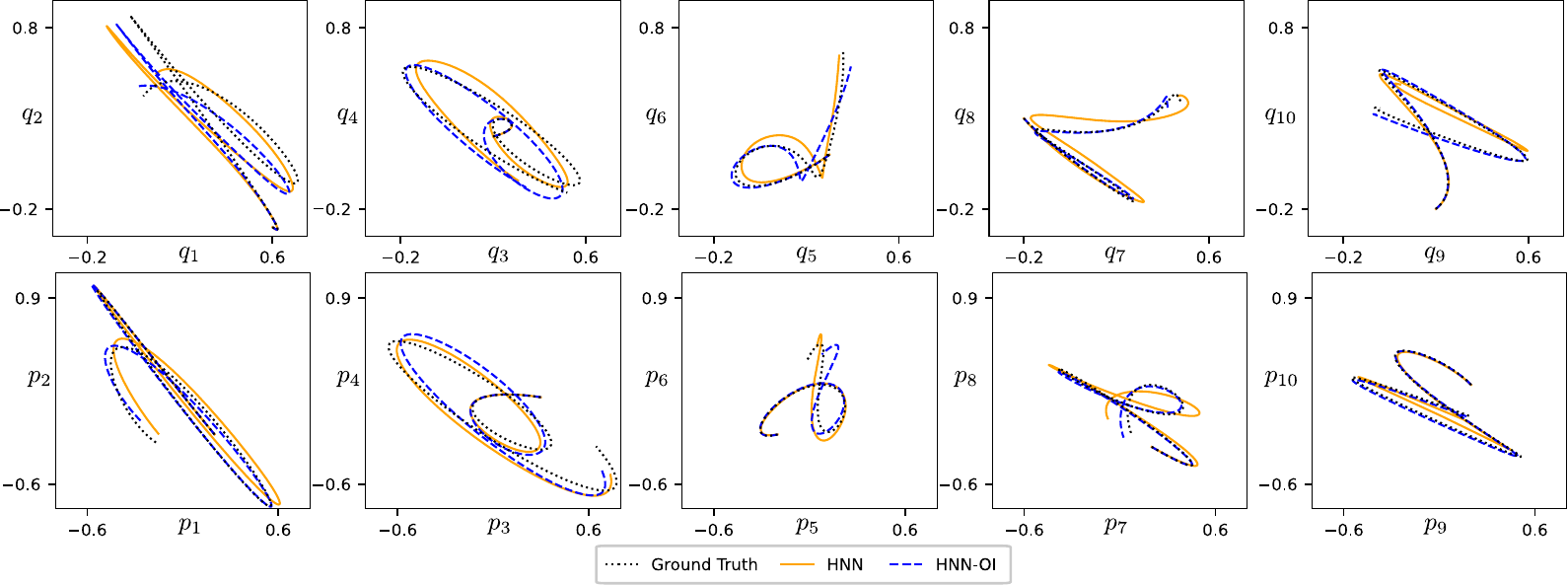}
    \caption{The trajectory of the Coupled Oscillator system with $n=10$ was obtained by integrating the ground truth equations of motion (in black, dotted line), the vector field from the HNN-OI (blue, dashed line), and the vector field from the HNN (orange, solid line). The dynamics in selected planes are shown in each plot.}
    \label{fig:CO_combine}
\end{figure*}

Fig.~\ref{fig:CO_time} and Fig.~\ref{fig:CO_combine} report the dynamics of the Coupled Oscillator system. We observe that generally, the HNN-OI predicts the dynamics of the Coupled Oscillator system better than the baseline HNN. The cumulative root mean squared error between the position of the HNN and the ground truth at every time step is 0.686. Between the HNN-OI and the ground truth, it is only 0.468. The difference between the trajectory of the HNN and HNN-OI from the ground truth increases with time. This is because the error in predicting the trajectory accumulates with every integration step. The root mean squared error between the final position of the HNN and the ground truth is 1.71. The root mean squared error between the final position of the HNN-OI and the ground truth is 0.575. Lastly, we observe that the HNN-OI conserves the Hamiltonian better than the HNN. The cumulative root mean squared error between the Hamiltonian given by the HNN and the ground truth Hamiltonian at every time step is 1.33. Between the HNN-OI and the ground truth, it is only 0.561. Therefore, the HNN-OI is significantly better, and the model of choice, for predicting the dynamics of the high dimensional Coupled Oscillator system.

\section{Conclusion} \label{sec:conclusion}

We propose \textit{separable HNNs} that embed additive separability using three modes of biases: observational bias, learning bias, and inductive bias. An observational bias is embedded by training on newly generated data that embody separability. A learning bias is embedded through the loss function of the HNN. An inductive bias is embedded by conjoined multilayer perceptrons in the HNN. Each proposed separable HNN may embed one or more biases. 
The proposed separable HNNs show high performance in regressing additively separable Hamiltonians and their vector fields. They leverage additive separability to avoid artificial complexity between input variables. 
The HNN-OI is the best as it outperforms the baseline in the regression tasks and has the smallest trade-off in training time. The HNN-OI also predicts the dynamics of the chaotic Henon Heiles system and the high dimensional Coupled Oscillator system more accurately than the baseline HNN.

As machine learning develops in the context of traditional science and engineering fields, more physical concepts and properties can be embedded in machine learning models, via observational, learning, or inductive biases for improved modeling and interpretation of dynamical systems. Additive separability is one such property, although generally, other properties such as symmetry, periodicity, and continuity can be embedded. We leave the investigation of these other properties for modeling and interpretation of dynamical systems for future work.

\section*{Acknowledgements}
This research is supported by Singapore Ministry of Education, grants MOE-T2EP50120-0019 and MoE-T1251RES2302, and by the National Research Foundation, Prime Minister’s Office, Singapore, under its Campus for Research Excellence and Technological Enterprise (CREATE) programme as part of the programme Descartes.

\section*{Appendix}

\appendix\subsection{Optimal architecture of the last layer.} \label{sec:lastlayer} The HNN-I is also trained and evaluated for five possible implementations of the summation layer:
\begin{itemize}
    \item L0: a simple summation
    \begin{eqnarray}
        \hat{H}_{IB}(q,p;w) = \hat{T}(q;w_1) + \hat{V}(p;w_2).
    \end{eqnarray}
    \item L1: a linear layer with fixed and equal weights, and no bias, 
    \begin{eqnarray}
        \hat{H}_{IB}(q,p;w) = \begin{bmatrix}
        \hat{T}(q;w_1) ,& \hat{V}(p;w_2)  
    \end{bmatrix} \times \begin{bmatrix}
        1 \\
        1  
    \end{bmatrix}.
    \end{eqnarray}
    L0 and L1 should have the same $E_H$ and $E_V$ but may differ in efficiency. 
    \item L2: a linear layer with fixed and equal weights, and a trainable bias
    \begin{eqnarray}
        &\hat{H}&_{IB}(q,p;w) \nonumber \\
        &\quad& = \begin{bmatrix}
        \hat{T}(q;w_1),  \hat{V}(p;w_2)  
    \end{bmatrix} \times \begin{bmatrix}
        1 \\
        1  
    \end{bmatrix} + \beta_{IB},
    \end{eqnarray}  
    where $\beta_{IB} \in \mathbb{R}$ is the trainable bias.
    \item L3: a linear layer with trainable weights, and no bias
    \begin{eqnarray}
        \hat{H}_{IB}(q,p;w) = \begin{bmatrix}
        \hat{T}(q;w_1), & \hat{V}(p;w_2)  
    \end{bmatrix} \times \begin{bmatrix}
        \alpha_{T} \\
        \alpha_{V}  
    \end{bmatrix},
    \end{eqnarray} 
    where $\alpha_{T} \in \mathbb{R}$ and $\alpha_{V} \in \mathbb{R}$ are the trainable weights.
    \item L4: a linear layer with trainable weights and bias
    \begin{eqnarray}
        &\hat{H}&_{IB}(q,p;w) \nonumber \\
        &\quad& = \begin{bmatrix}
        \hat{T}(q;w_1),  \hat{V}(p;w_2)  
    \end{bmatrix} \times \begin{bmatrix}
        \alpha_{T} \\
        \alpha_{V}  
    \end{bmatrix} + \beta_{IB},
    \end{eqnarray} 
    where $\alpha_{T} \in \mathbb{R}$ and $\alpha_{V} \in \mathbb{R}$ are the trainable weights and $\beta_{IB} \in \mathbb{R}$ is the trainable bias.
\end{itemize}

\begin{figure*}
    \centering
    \includegraphics[width = \textwidth]{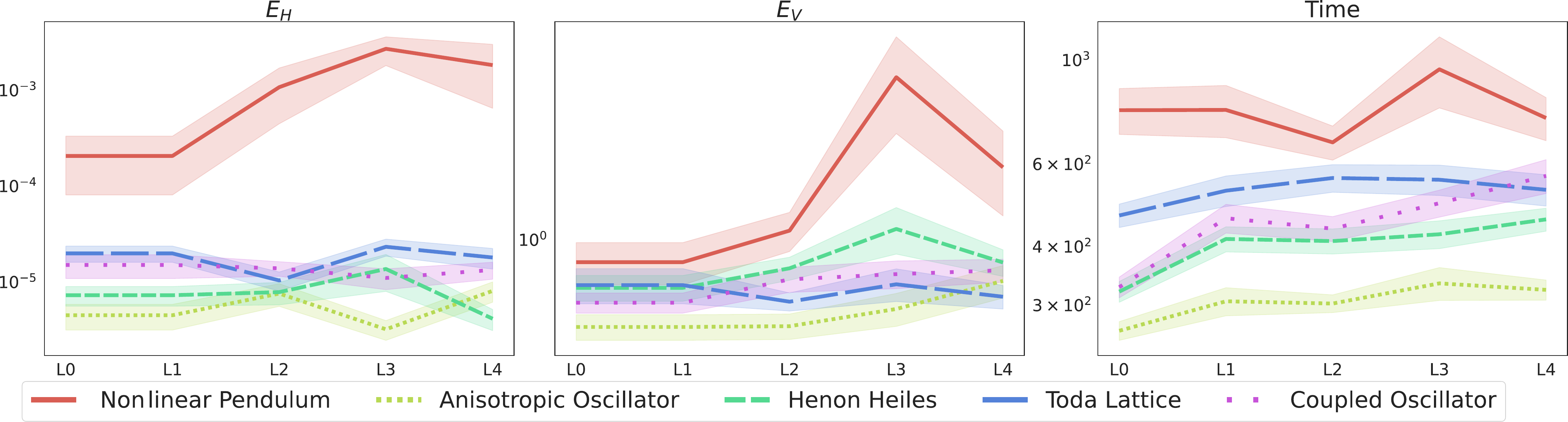}
    \caption{Performance of the HNN-I with five possible implementations of the summation layer for the Nonlinear Pendulum (red, solid line), Anisotropic Oscillator (yellow, densely dotted line), Henon Heiles (green, loosely dashed line), Toda Lattice (blue, densely dashed line), and Coupled Oscillator (purple, loosely dotted line) systems. Left: $E_H$ and standard error. Center: $E_V$ and standard error. Right: Time taken until convergence in seconds and standard error.}
    \label{fig:inductive_error_lastlayer}
\end{figure*}
The last column in Table~\ref{tab:nnsetup} shows the number of parameters of the HNN-I with L0 and L1 implementations. The L2, L3, and L4 implementations have one, two, and three additional parameters respectively. 
All other details of the setup, training, and evaluation follow from the description above in Sec.~\ref{sec:exp}.

We report the $E_H$, $E_V$, and the training time for the last layer of the HNN-I in Fig.~\ref{fig:inductive_error_lastlayer}. Generally, L0 and L1 have the lowest $E_H$ and $E_V$. However, L0 is consistently faster than L1.  
As a result, L0 is the preferred implementation of the HNN-I. 

\subsection{Interpreting the HNNs with Inductive Bias}~\label{sec:expKEPE}
We focus on empirically interpreting the conjoined neural networks within the HNNs with an inductive bias. The purpose of this empirical interpretation is to interpret the two conjoined neural networks separately, to disentangle and identify the kinetic and potential energy of the Hamiltonian system.

The HNN-I predicts the Hamiltonian following Eq.\eqref{eqn:HNNI}. Therefore, the two conjoined neural networks predict the kinetic and potential energy. The kinetic and potential energy for the HNN-I, HNN-LI, HNN-OI, and HNN-OLI are evaluated. 
The mean squared error for regressing the kinetic and potential energy of the various Hamiltonian systems are
\begin{eqnarray} 
    E_{KE} =\frac{1}{K} \sum^{K}_{k=1} \left(\hat{V}(p_k; w)-V(p_k)\right)^2,  \label{eqn:KEerror} \\
    E_{PE} =\frac{1}{K} \sum^{K}_{k=1} \left(\hat{T}(q_k; w)-T(q_k)\right)^2,  \label{eqn:PEerror} 
\end{eqnarray}
where $\hat{V}(p_k; w)$ and $V(p_k)$  are the predicted and true kinetic energy, respectively. $\hat{T}(q_k; w)$ and $T(q_k)$ are the predicted and true potential energy, respectively. $K$ is the size of the test batch of $10^{2n}$ evenly spaced samples.

For the purpose of evaluation, we arbitrarily pin the potential and kinetic energies to $T(0) = 0$ and $V(0)=0$. 

\begin{figure}[h]
    \centering
    \includegraphics[width = 0.45\textwidth]{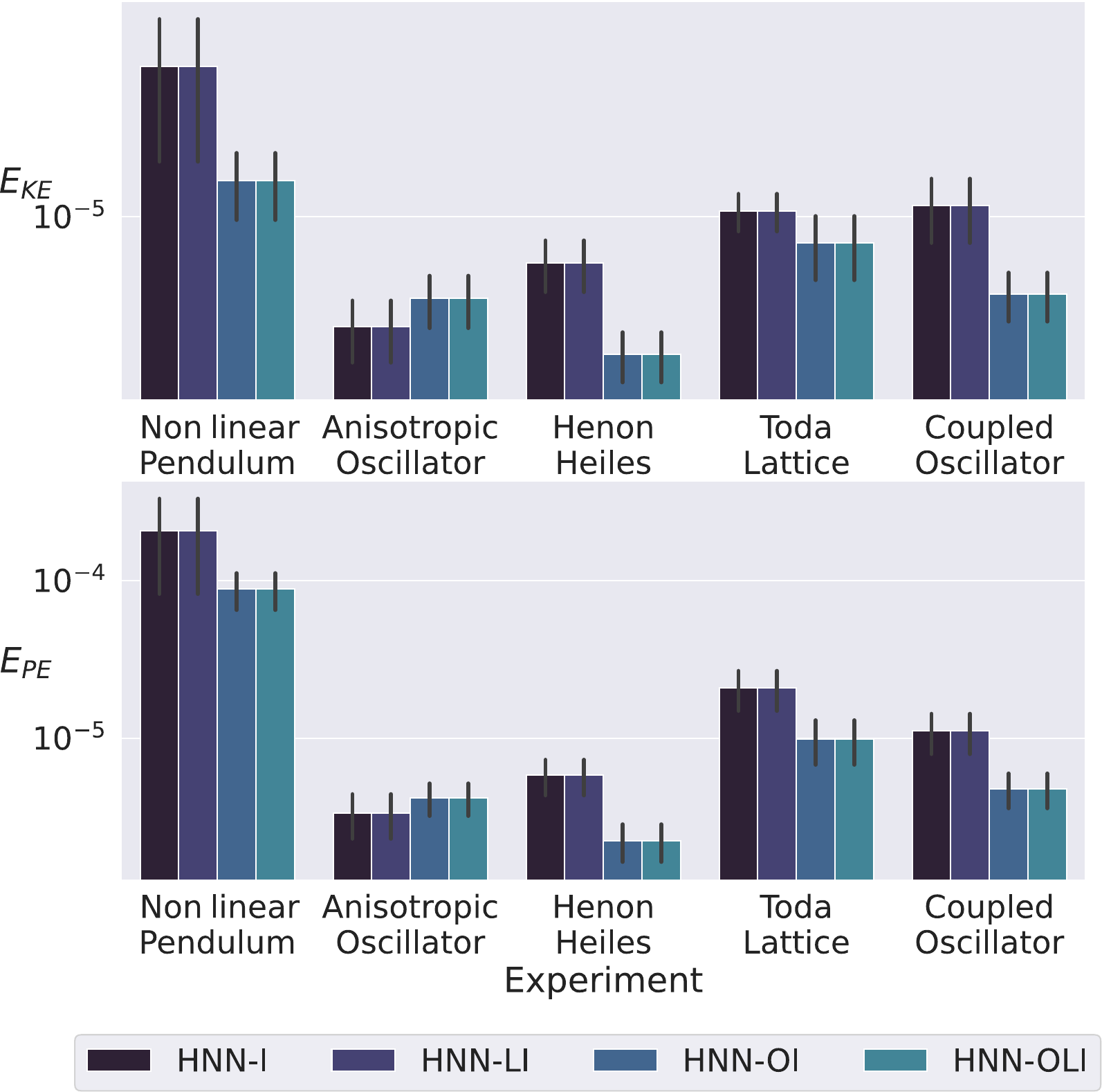}
    \caption{Interpreting the HNN-I, HNN-LI, HNN-OI, and HNN-OLI. Top: $E_{KE}$ and the standard error. Bottom: $E_{PE}$ and standard error.The bar plots in each group are presented in the order shown in the legend.}
    \label{fig:KError}
\end{figure}

We report $E_{KE}$ and $E_{PE}$ for the separable and mechanical Hamiltonian systems in Fig.~\ref{fig:KError}. We note the lack of a baseline, as the baseline HNN does not allow interpretation of the kinetic and potential energies of the regressed Hamiltonian. Furthermore, the learning bias becomes redundant when paired with an inductive bias. Therefore, the HNN-I and HNN-LI, and the HNN-OI and HNN-OLI have the same $E_{KE}$ and $E_{PE}$. Generally, the HNN-OI and HNN-OLI have the lowest $E_{KE}$ and $E_{PE}$ and are the models of choice for interpreting the kinetic and potential energy of a Hamiltonian system. 

\bibliography{apssamp}

\providecommand{\noopsort}[1]{}\providecommand{\singleletter}[1]{#1}%
\begin{thebibliography}{23}%
\makeatletter
\providecommand \@ifxundefined [1]{%
 \@ifx{#1\undefined}
}%
\providecommand \@ifnum [1]{%
 \ifnum #1\expandafter \@firstoftwo
 \else \expandafter \@secondoftwo
 \fi
}%
\providecommand \@ifx [1]{%
 \ifx #1\expandafter \@firstoftwo
 \else \expandafter \@secondoftwo
 \fi
}%
\providecommand \natexlab [1]{#1}%
\providecommand \enquote  [1]{``#1''}%
\providecommand \bibnamefont  [1]{#1}%
\providecommand \bibfnamefont [1]{#1}%
\providecommand \citenamefont [1]{#1}%
\providecommand \href@noop [0]{\@secondoftwo}%
\providecommand \href [0]{\begingroup \@sanitize@url \@href}%
\providecommand \@href[1]{\@@startlink{#1}\@@href}%
\providecommand \@@href[1]{\endgroup#1\@@endlink}%
\providecommand \@sanitize@url [0]{\catcode `\\12\catcode `\$12\catcode `\&12\catcode `\#12\catcode `\^12\catcode `\_12\catcode `\%12\relax}%
\providecommand \@@startlink[1]{}%
\providecommand \@@endlink[0]{}%
\providecommand \url  [0]{\begingroup\@sanitize@url \@url }%
\providecommand \@url [1]{\endgroup\@href {#1}{\urlprefix }}%
\providecommand \urlprefix  [0]{URL }%
\providecommand \Eprint [0]{\href }%
\providecommand \doibase [0]{https://doi.org/}%
\providecommand \selectlanguage [0]{\@gobble}%
\providecommand \bibinfo  [0]{\@secondoftwo}%
\providecommand \bibfield  [0]{\@secondoftwo}%
\providecommand \translation [1]{[#1]}%
\providecommand \BibitemOpen [0]{}%
\providecommand \bibitemStop [0]{}%
\providecommand \bibitemNoStop [0]{.\EOS\space}%
\providecommand \EOS [0]{\spacefactor3000\relax}%
\providecommand \BibitemShut  [1]{\csname bibitem#1\endcsname}%
\let\auto@bib@innerbib\@empty
\bibitem [{\citenamefont {Meyer}\ and\ \citenamefont {Hall}(1992)}]{Meyer1992}%
  \BibitemOpen
  \bibfield  {author} {\bibinfo {author} {\bibfnamefont {K.~R.}\ \bibnamefont {Meyer}}\ and\ \bibinfo {author} {\bibfnamefont {G.~R.}\ \bibnamefont {Hall}},\ }\bibinfo {title} {Hamiltonian differential equations and the n-body problem},\ in\ \href {https://doi.org/10.1007/978-1-4757-4073-8\_1} {\emph {\bibinfo {booktitle} {Introduction to Hamiltonian Dynamical Systems and the N-Body Problem}}}\ (\bibinfo  {publisher} {Springer New York},\ \bibinfo {address} {New York, NY},\ \bibinfo {year} {1992})\ pp.\ \bibinfo {pages} {1--32}\BibitemShut {NoStop}%
\bibitem [{\citenamefont {Mattheakis}\ \emph {et~al.}(2022)\citenamefont {Mattheakis}, \citenamefont {Sondak}, \citenamefont {Dogra},\ and\ \citenamefont {Protopapas}}]{Marios2022}%
  \BibitemOpen
  \bibfield  {author} {\bibinfo {author} {\bibfnamefont {M.}~\bibnamefont {Mattheakis}}, \bibinfo {author} {\bibfnamefont {D.}~\bibnamefont {Sondak}}, \bibinfo {author} {\bibfnamefont {A.~S.}\ \bibnamefont {Dogra}},\ and\ \bibinfo {author} {\bibfnamefont {P.}~\bibnamefont {Protopapas}},\ }\bibfield  {title} {\bibinfo {title} {Hamiltonian neural networks for solving equations of motion},\ }\href {https://doi.org/10.1103/PhysRevE.105.065305} {\bibfield  {journal} {\bibinfo  {journal} {Phys. Rev. E}\ }\textbf {\bibinfo {volume} {105}},\ \bibinfo {pages} {065305} (\bibinfo {year} {2022})}\BibitemShut {NoStop}%
\bibitem [{\citenamefont {Bertalan}\ \emph {et~al.}(2019)\citenamefont {Bertalan}, \citenamefont {Dietrich}, \citenamefont {Mezi{\'{c} }},\ and\ \citenamefont {Kevrekidis}}]{bertalan_2019}%
  \BibitemOpen
  \bibfield  {author} {\bibinfo {author} {\bibfnamefont {T.}~\bibnamefont {Bertalan}}, \bibinfo {author} {\bibfnamefont {F.}~\bibnamefont {Dietrich}}, \bibinfo {author} {\bibfnamefont {I.}~\bibnamefont {Mezi{\'{c} }}},\ and\ \bibinfo {author} {\bibfnamefont {I.~G.}\ \bibnamefont {Kevrekidis}},\ }\bibfield  {title} {\bibinfo {title} {On learning hamiltonian systems from data},\ }\href {https://doi.org/10.1063/1.5128231} {\bibfield  {journal} {\bibinfo  {journal} {Chaos: An Interdisciplinary Journal of Nonlinear Science}\ }\textbf {\bibinfo {volume} {29}},\ \bibinfo {pages} {121107} (\bibinfo {year} {2019})}\BibitemShut {NoStop}%
\bibitem [{\citenamefont {Greydanus}\ \emph {et~al.}(2019)\citenamefont {Greydanus}, \citenamefont {Dzamba},\ and\ \citenamefont {Yosinski}}]{Greydanus_hamiltoniannn_2019}%
  \BibitemOpen
  \bibfield  {author} {\bibinfo {author} {\bibfnamefont {S.}~\bibnamefont {Greydanus}}, \bibinfo {author} {\bibfnamefont {M.}~\bibnamefont {Dzamba}},\ and\ \bibinfo {author} {\bibfnamefont {J.}~\bibnamefont {Yosinski}},\ }\bibfield  {title} {\bibinfo {title} {Hamiltonian neural networks},\ }in\ \href {https://proceedings.neurips.cc/paper/2019/file/26cd8ecadce0d4efd6cc8a8725cbd1f8-Paper.pdf} {\emph {\bibinfo {booktitle} {Advances in Neural Information Processing Systems}}},\ Vol.~\bibinfo {volume} {32},\ \bibinfo {editor} {edited by\ \bibinfo {editor} {\bibfnamefont {H.}~\bibnamefont {Wallach}}, \bibinfo {editor} {\bibfnamefont {H.}~\bibnamefont {Larochelle}}, \bibinfo {editor} {\bibfnamefont {A.}~\bibnamefont {Beygelzimer}}, \bibinfo {editor} {\bibfnamefont {F.}~\bibnamefont {d\textquotesingle Alch\'{e}-Buc}}, \bibinfo {editor} {\bibfnamefont {E.}~\bibnamefont {Fox}},\ and\ \bibinfo {editor} {\bibfnamefont {R.}~\bibnamefont {Garnett}}}\ (\bibinfo  {publisher} {Curran Associates, Inc.},\ \bibinfo {year}
  {2019})\BibitemShut {NoStop}%
\bibitem [{\citenamefont {Desai}\ \emph {et~al.}(2021)\citenamefont {Desai}, \citenamefont {Mattheakis}, \citenamefont {Sondak}, \citenamefont {Protopapas},\ and\ \citenamefont {Roberts}}]{Desai2021}%
  \BibitemOpen
  \bibfield  {author} {\bibinfo {author} {\bibfnamefont {S.~A.}\ \bibnamefont {Desai}}, \bibinfo {author} {\bibfnamefont {M.}~\bibnamefont {Mattheakis}}, \bibinfo {author} {\bibfnamefont {D.}~\bibnamefont {Sondak}}, \bibinfo {author} {\bibfnamefont {P.}~\bibnamefont {Protopapas}},\ and\ \bibinfo {author} {\bibfnamefont {S.~J.}\ \bibnamefont {Roberts}},\ }\bibfield  {title} {\bibinfo {title} {Port-hamiltonian neural networks for learning explicit time-dependent dynamical systems},\ }\href {https://doi.org/10.1103/PhysRevE.104.034312} {\bibfield  {journal} {\bibinfo  {journal} {Phys. Rev. E}\ }\textbf {\bibinfo {volume} {104}},\ \bibinfo {pages} {034312} (\bibinfo {year} {2021})}\BibitemShut {NoStop}%
\bibitem [{\citenamefont {Zhang}\ \emph {et~al.}(2021)\citenamefont {Zhang}, \citenamefont {Fan}, \citenamefont {Wang},\ and\ \citenamefont {Wang}}]{Zhang2021}%
  \BibitemOpen
  \bibfield  {author} {\bibinfo {author} {\bibfnamefont {H.}~\bibnamefont {Zhang}}, \bibinfo {author} {\bibfnamefont {H.}~\bibnamefont {Fan}}, \bibinfo {author} {\bibfnamefont {L.}~\bibnamefont {Wang}},\ and\ \bibinfo {author} {\bibfnamefont {X.}~\bibnamefont {Wang}},\ }\bibfield  {title} {\bibinfo {title} {Learning hamiltonian dynamics with reservoir computing},\ }\href {https://doi.org/10.1103/PhysRevE.104.024205} {\bibfield  {journal} {\bibinfo  {journal} {Phys. Rev. E}\ }\textbf {\bibinfo {volume} {104}},\ \bibinfo {pages} {024205} (\bibinfo {year} {2021})}\BibitemShut {NoStop}%
\bibitem [{\citenamefont {Chen}\ \emph {et~al.}(2020)\citenamefont {Chen}, \citenamefont {Zhang}, \citenamefont {Arjovsky},\ and\ \citenamefont {Bottou}}]{Chen_symplectic2020}%
  \BibitemOpen
  \bibfield  {author} {\bibinfo {author} {\bibfnamefont {Z.}~\bibnamefont {Chen}}, \bibinfo {author} {\bibfnamefont {J.}~\bibnamefont {Zhang}}, \bibinfo {author} {\bibfnamefont {M.}~\bibnamefont {Arjovsky}},\ and\ \bibinfo {author} {\bibfnamefont {L.}~\bibnamefont {Bottou}},\ }\bibfield  {title} {\bibinfo {title} {Symplectic recurrent neural networks},\ }in\ \href {https://openreview.net/forum?id=BkgYPREtPr} {\emph {\bibinfo {booktitle} {International Conference on Learning Representations}}}\ (\bibinfo {year} {2020})\BibitemShut {NoStop}%
\bibitem [{\citenamefont {Toth}\ \emph {et~al.}(2020)\citenamefont {Toth}, \citenamefont {Rezende}, \citenamefont {Jaegle}, \citenamefont {Racanière}, \citenamefont {Botev},\ and\ \citenamefont {Higgins}}]{TothHGN}%
  \BibitemOpen
  \bibfield  {author} {\bibinfo {author} {\bibfnamefont {P.}~\bibnamefont {Toth}}, \bibinfo {author} {\bibfnamefont {D.~J.}\ \bibnamefont {Rezende}}, \bibinfo {author} {\bibfnamefont {A.}~\bibnamefont {Jaegle}}, \bibinfo {author} {\bibfnamefont {S.}~\bibnamefont {Racanière}}, \bibinfo {author} {\bibfnamefont {A.}~\bibnamefont {Botev}},\ and\ \bibinfo {author} {\bibfnamefont {I.}~\bibnamefont {Higgins}},\ }\bibfield  {title} {\bibinfo {title} {Hamiltonian generative networks},\ }in\ \href {https://openreview.net/forum?id=HJenn6VFvB} {\emph {\bibinfo {booktitle} {International Conference on Learning Representations}}}\ (\bibinfo {year} {2020})\BibitemShut {NoStop}%
\bibitem [{\citenamefont {Zhong}\ \emph {et~al.}(2020)\citenamefont {Zhong}, \citenamefont {Dey},\ and\ \citenamefont {Chakraborty}}]{Zhong2019SymplecticOL}%
  \BibitemOpen
  \bibfield  {author} {\bibinfo {author} {\bibfnamefont {Y.~D.}\ \bibnamefont {Zhong}}, \bibinfo {author} {\bibfnamefont {B.}~\bibnamefont {Dey}},\ and\ \bibinfo {author} {\bibfnamefont {A.}~\bibnamefont {Chakraborty}},\ }\bibfield  {title} {\bibinfo {title} {Symplectic ode-net: Learning hamiltonian dynamics with control},\ }in\ \href {https://openreview.net/forum?id=ryxmb1rKDS} {\emph {\bibinfo {booktitle} {International Conference on Learning Representations}}}\ (\bibinfo {year} {2020})\BibitemShut {NoStop}%
\bibitem [{\citenamefont {Karniadakis}\ \emph {et~al.}(2021)\citenamefont {Karniadakis}, \citenamefont {Kevrekidis}, \citenamefont {Lu}, \citenamefont {Perdikaris}, \citenamefont {Wang},\ and\ \citenamefont {Yang}}]{Karniadakis2021}%
  \BibitemOpen
  \bibfield  {author} {\bibinfo {author} {\bibfnamefont {G.~E.}\ \bibnamefont {Karniadakis}}, \bibinfo {author} {\bibfnamefont {I.~G.}\ \bibnamefont {Kevrekidis}}, \bibinfo {author} {\bibfnamefont {L.}~\bibnamefont {Lu}}, \bibinfo {author} {\bibfnamefont {P.}~\bibnamefont {Perdikaris}}, \bibinfo {author} {\bibfnamefont {S.}~\bibnamefont {Wang}},\ and\ \bibinfo {author} {\bibfnamefont {L.}~\bibnamefont {Yang}},\ }\bibfield  {title} {\bibinfo {title} {Physics-informed machine learning},\ }\href {https://doi.org/10.1038/s42254-021-00314-5} {\bibfield  {journal} {\bibinfo  {journal} {Nature Reviews Physics}\ }\textbf {\bibinfo {volume} {3}},\ \bibinfo {pages} {422} (\bibinfo {year} {2021})}\BibitemShut {NoStop}%
\bibitem [{\citenamefont {Khoo}\ \emph {et~al.}(2022)\citenamefont {Khoo}, \citenamefont {Zhang},\ and\ \citenamefont {Bressan}}]{khoo_22}%
  \BibitemOpen
  \bibfield  {author} {\bibinfo {author} {\bibfnamefont {Z.-Y.}\ \bibnamefont {Khoo}}, \bibinfo {author} {\bibfnamefont {D.}~\bibnamefont {Zhang}},\ and\ \bibinfo {author} {\bibfnamefont {S.}~\bibnamefont {Bressan}},\ }\bibfield  {title} {\bibinfo {title} {What’s next? predicting hamiltonian dynamics from discrete observations of a vector field},\ }in\ \href@noop {} {\emph {\bibinfo {booktitle} {Database and Expert Systems Applications: 33rd International Conference, DEXA 2022, Vienna, Austria, August 22–24, 2022, Proceedings, Part II}}}\ (\bibinfo  {publisher} {Springer-Verlag},\ \bibinfo {address} {Berlin, Heidelberg},\ \bibinfo {year} {2022})\ p.\ \bibinfo {pages} {297–302}\BibitemShut {NoStop}%
\bibitem [{\citenamefont {Toda}(1967)}]{Toda1967}%
  \BibitemOpen
  \bibfield  {author} {\bibinfo {author} {\bibfnamefont {M.}~\bibnamefont {Toda}},\ }\bibfield  {title} {\bibinfo {title} {Vibration of a chain with nonlinear interaction},\ }\href {https://doi.org/10.1143/JPSJ.22.431} {\bibfield  {journal} {\bibinfo  {journal} {Journal of the Physical Society of Japan}\ }\textbf {\bibinfo {volume} {22}},\ \bibinfo {pages} {431} (\bibinfo {year} {1967})},\ \Eprint {https://arxiv.org/abs/https://doi.org/10.1143/JPSJ.22.431} {https://doi.org/10.1143/JPSJ.22.431} \BibitemShut {NoStop}%
\bibitem [{\citenamefont {Ford}\ \emph {et~al.}(1973)\citenamefont {Ford}, \citenamefont {Stoddard},\ and\ \citenamefont {Turner}}]{Ford1973}%
  \BibitemOpen
  \bibfield  {author} {\bibinfo {author} {\bibfnamefont {J.}~\bibnamefont {Ford}}, \bibinfo {author} {\bibfnamefont {S.~D.}\ \bibnamefont {Stoddard}},\ and\ \bibinfo {author} {\bibfnamefont {J.~S.}\ \bibnamefont {Turner}},\ }\bibfield  {title} {\bibinfo {title} {{On the Integrability of the Toda Lattice}},\ }\href {https://doi.org/10.1143/PTP.50.1547} {\bibfield  {journal} {\bibinfo  {journal} {Progress of Theoretical Physics}\ }\textbf {\bibinfo {volume} {50}},\ \bibinfo {pages} {1547} (\bibinfo {year} {1973})},\ \Eprint {https://arxiv.org/abs/https://academic.oup.com/ptp/article-pdf/50/5/1547/5206486/50-5-1547.pdf} {https://academic.oup.com/ptp/article-pdf/50/5/1547/5206486/50-5-1547.pdf} \BibitemShut {NoStop}%
\bibitem [{\citenamefont {{Henon}}\ and\ \citenamefont {{Heiles}}(1964)}]{henonheiles1964}%
  \BibitemOpen
  \bibfield  {author} {\bibinfo {author} {\bibfnamefont {M.}~\bibnamefont {{Henon}}}\ and\ \bibinfo {author} {\bibfnamefont {C.}~\bibnamefont {{Heiles}}},\ }\bibfield  {title} {\bibinfo {title} {{The applicability of the third integral of motion: Some numerical experiments}},\ }\href {https://doi.org/10.1086/109234} {\bibfield  {journal} {\bibinfo  {journal} {The Astronomical Journal}\ }\textbf {\bibinfo {volume} {69}},\ \bibinfo {pages} {73} (\bibinfo {year} {1964})}\BibitemShut {NoStop}%
\bibitem [{\citenamefont {Gruver}\ \emph {et~al.}(2022)\citenamefont {Gruver}, \citenamefont {Finzi}, \citenamefont {Stanton},\ and\ \citenamefont {Wilson}}]{gruver2022deconstructing}%
  \BibitemOpen
  \bibfield  {author} {\bibinfo {author} {\bibfnamefont {N.}~\bibnamefont {Gruver}}, \bibinfo {author} {\bibfnamefont {M.~A.}\ \bibnamefont {Finzi}}, \bibinfo {author} {\bibfnamefont {S.~D.}\ \bibnamefont {Stanton}},\ and\ \bibinfo {author} {\bibfnamefont {A.~G.}\ \bibnamefont {Wilson}},\ }\bibfield  {title} {\bibinfo {title} {Deconstructing the inductive biases of hamiltonian neural networks},\ }in\ \href {https://openreview.net/forum?id=EDeVYpT42oS} {\emph {\bibinfo {booktitle} {International Conference on Learning Representations}}}\ (\bibinfo {year} {2022})\BibitemShut {NoStop}%
\bibitem [{\citenamefont {Udrescu}\ and\ \citenamefont {Tegmark}(2020)}]{Udrescu_2020}%
  \BibitemOpen
  \bibfield  {author} {\bibinfo {author} {\bibfnamefont {S.-M.}\ \bibnamefont {Udrescu}}\ and\ \bibinfo {author} {\bibfnamefont {M.}~\bibnamefont {Tegmark}},\ }\bibfield  {title} {\bibinfo {title} {{AI Feynman}: A physics-inspired method for symbolic regression},\ }\bibfield  {journal} {\bibinfo  {journal} {Science Advances}\ }\textbf {\bibinfo {volume} {6}},\ \href {https://doi.org/10.1126/sciadv.aay2631} {10.1126/sciadv.aay2631} (\bibinfo {year} {2020}),\ \Eprint {https://arxiv.org/abs/https://advances.sciencemag.org/content/6/16/eaay2631.full.pdf} {https://advances.sciencemag.org/content/6/16/eaay2631.full.pdf} \BibitemShut {NoStop}%
\bibitem [{\citenamefont {Bellenot}()}]{bellenot_addsep}%
  \BibitemOpen
  \bibfield  {author} {\bibinfo {author} {\bibfnamefont {S.~F.}\ \bibnamefont {Bellenot}},\ }\href {https://www.math.fsu.edu/~bellenot/class/s05/cal3/proj/project.pdf} {\bibinfo {title} {Additively separable functions of the form f(x,y)=f(x)+g(y)}}\BibitemShut {NoStop}%
\bibitem [{\citenamefont {Prechelt}(1998)}]{Prechelt1998}%
  \BibitemOpen
  \bibfield  {author} {\bibinfo {author} {\bibfnamefont {L.}~\bibnamefont {Prechelt}},\ }\bibfield  {title} {\bibinfo {title} {Early stopping-but when?},\ }in\ \href@noop {} {\emph {\bibinfo {booktitle} {Neural Networks: Tricks of the Trade, This Book is an Outgrowth of a 1996 NIPS Workshop}}}\ (\bibinfo  {publisher} {Springer-Verlag},\ \bibinfo {address} {Berlin, Heidelberg},\ \bibinfo {year} {1998})\ p.\ \bibinfo {pages} {55–69}\BibitemShut {NoStop}%
\bibitem [{\citenamefont {Kingma}\ and\ \citenamefont {Ba}(2014)}]{kingma2014adam}%
  \BibitemOpen
  \bibfield  {author} {\bibinfo {author} {\bibfnamefont {D.~P.}\ \bibnamefont {Kingma}}\ and\ \bibinfo {author} {\bibfnamefont {J.}~\bibnamefont {Ba}},\ }\bibfield  {title} {\bibinfo {title} {Adam: A method for stochastic optimization},\ }\href@noop {} {\bibfield  {journal} {\bibinfo  {journal} {arXiv preprint arXiv:1412.6980}\ } (\bibinfo {year} {2014})}\BibitemShut {NoStop}%
\bibitem [{\citenamefont {Paszke}\ \emph {et~al.}(2017)\citenamefont {Paszke}, \citenamefont {Gross}, \citenamefont {Chintala}, \citenamefont {Chanan}, \citenamefont {Yang}, \citenamefont {DeVito}, \citenamefont {Lin}, \citenamefont {Desmaison}, \citenamefont {Antiga},\ and\ \citenamefont {Lerer}}]{pytorch-autodiff}%
  \BibitemOpen
  \bibfield  {author} {\bibinfo {author} {\bibfnamefont {A.}~\bibnamefont {Paszke}}, \bibinfo {author} {\bibfnamefont {S.}~\bibnamefont {Gross}}, \bibinfo {author} {\bibfnamefont {S.}~\bibnamefont {Chintala}}, \bibinfo {author} {\bibfnamefont {G.}~\bibnamefont {Chanan}}, \bibinfo {author} {\bibfnamefont {E.}~\bibnamefont {Yang}}, \bibinfo {author} {\bibfnamefont {Z.}~\bibnamefont {DeVito}}, \bibinfo {author} {\bibfnamefont {Z.}~\bibnamefont {Lin}}, \bibinfo {author} {\bibfnamefont {A.}~\bibnamefont {Desmaison}}, \bibinfo {author} {\bibfnamefont {L.}~\bibnamefont {Antiga}},\ and\ \bibinfo {author} {\bibfnamefont {A.}~\bibnamefont {Lerer}},\ }\href@noop {} {\bibinfo {title} {Automatic differentiation in pytorch}} (\bibinfo {year} {2017})\BibitemShut {NoStop}%
\bibitem [{\citenamefont {Bradbury}\ \emph {et~al.}(2023)\citenamefont {Bradbury}, \citenamefont {Frostig}, \citenamefont {Hawkins}, \citenamefont {Johnson}, \citenamefont {Leary}, \citenamefont {Maclaurin}, \citenamefont {Necula}, \citenamefont {Paszke}, \citenamefont {Vander{P}las}, \citenamefont {Wanderman-{M}ilne},\ and\ \citenamefont {Zhang}}]{jax2018github}%
  \BibitemOpen
  \bibfield  {author} {\bibinfo {author} {\bibfnamefont {J.}~\bibnamefont {Bradbury}}, \bibinfo {author} {\bibfnamefont {R.}~\bibnamefont {Frostig}}, \bibinfo {author} {\bibfnamefont {P.}~\bibnamefont {Hawkins}}, \bibinfo {author} {\bibfnamefont {M.~J.}\ \bibnamefont {Johnson}}, \bibinfo {author} {\bibfnamefont {C.}~\bibnamefont {Leary}}, \bibinfo {author} {\bibfnamefont {D.}~\bibnamefont {Maclaurin}}, \bibinfo {author} {\bibfnamefont {G.}~\bibnamefont {Necula}}, \bibinfo {author} {\bibfnamefont {A.}~\bibnamefont {Paszke}}, \bibinfo {author} {\bibfnamefont {J.}~\bibnamefont {Vander{P}las}}, \bibinfo {author} {\bibfnamefont {S.}~\bibnamefont {Wanderman-{M}ilne}},\ and\ \bibinfo {author} {\bibfnamefont {Q.}~\bibnamefont {Zhang}},\ }\href {http://github.com/google/jax} {\bibinfo {title} {{JAX}: composable transformations of {P}ython+{N}um{P}y programs}} (\bibinfo {year} {2023})\BibitemShut {NoStop}%
\bibitem [{\citenamefont {Leimkuhler}\ and\ \citenamefont {Skeel}(1994)}]{LEIMKUHLER1994117}%
  \BibitemOpen
  \bibfield  {author} {\bibinfo {author} {\bibfnamefont {B.~J.}\ \bibnamefont {Leimkuhler}}\ and\ \bibinfo {author} {\bibfnamefont {R.~D.}\ \bibnamefont {Skeel}},\ }\bibfield  {title} {\bibinfo {title} {Symplectic numerical integrators in constrained hamiltonian systems},\ }\href {https://doi.org/https://doi.org/10.1006/jcph.1994.1085} {\bibfield  {journal} {\bibinfo  {journal} {Journal of Computational Physics}\ }\textbf {\bibinfo {volume} {112}},\ \bibinfo {pages} {117} (\bibinfo {year} {1994})}\BibitemShut {NoStop}%
\bibitem [{\citenamefont {Shevchenko}\ and\ \citenamefont {Mel'nikov}(2003)}]{Shevchenko2003}%
  \BibitemOpen
  \bibfield  {author} {\bibinfo {author} {\bibfnamefont {I.~I.}\ \bibnamefont {Shevchenko}}\ and\ \bibinfo {author} {\bibfnamefont {A.~V.}\ \bibnamefont {Mel'nikov}},\ }\bibfield  {title} {\bibinfo {title} {Lyapunov exponents in the h{\'e}non-heiles problem},\ }\href {https://doi.org/10.1134/1.1604412} {\bibfield  {journal} {\bibinfo  {journal} {Journal of Experimental and Theoretical Physics Letters}\ }\textbf {\bibinfo {volume} {77}},\ \bibinfo {pages} {642} (\bibinfo {year} {2003})}\BibitemShut {NoStop}%
\end{thebibliography}%

\end{document}